\documentclass[runningheads]{ECCV2024/llncs}

\usepackage{ECCV2024/eccv}

\usepackage{ECCV2024/eccvabbrv}
\usepackage{multicol}
\usepackage{multirow}
\usepackage{amsmath}
\usepackage{algorithmic}
\usepackage{algorithm}

\usepackage{graphicx}
\usepackage{booktabs}
\newcommand{\figref}[1]{Fig.~\ref{#1}}
\newcommand{\tabref}[1]{\cref{#1}}
\newcommand{\secref}[1]{Sec.~\ref{#1}}
\newcommand{\eqnref}[1]{\cref{#1}} 
\newcommand{\algref}[1]{Alg.~\ref{#1}}
\DeclareMathOperator*{\argmin}{arg\,min}
\let\titleold\title
\renewcommand{\title}[1]{\titleold{#1}\newcommand{\thetitle}{#1}}
\def\maketitlesupplementary
   {
   \newpage
       { \centering
        \Large
        \textbf{\thetitle}\\
        \vspace{0.5em}Supplementary Material \\
        \vspace{1.0em}}
   }

\newcommand{\mainpaperref}[1]{{\color{red}#1}}
\newcommand{\suppmatref}[1]{{\color{red}#1}}

\usepackage[accsupp]{axessibility}  %

\usepackage{hyperref}

\usepackage{orcidlink}

\begin{document}

\title{AugUndo: Scaling Up Augmentations for Monocular Depth Completion and Estimation}

\titlerunning{AugUndo: Scaling Up Augmentations for Depth Completion \& Estimation}

\author{
    Yangchao Wu\inst{1}$\orcidlink{0000-0002-6260-9693}$\index{Wu, Yangchao} \and Tian Yu Liu\inst{1}$\orcidlink{0000-0002-4834-051X}$\index{Liu, Tian Yu} \and Hyoungseob Park\inst{2}$\orcidlink{0000-0003-0787-2082}$\index{Park, Hyoungseob} \and \\
    Stefano Soatto\inst{1}$\orcidlink{0000-0003-2902-6362}$\index{Soatto, Stefano} \and Dong Lao\inst{1}$\orcidlink{0000-0001-9308-7085}$\index{Lao, Dong} \and Alex Wong\inst{2}$\orcidlink{0000-0002-3157-6016}$\index{Wong, Alex}  
}

\institute{
UCLA Vision Lab, Los Angeles, CA 90095, USA \\
\email{\{wuyangchao1997, tianyu, soatto, lao\}@cs.ucla.edu} \and
Yale Vision Lab, New Haven, CT 06511, USA \\
\email{\{hyoungseob.park, alex.wong\}@yale.edu} \\
}

\authorrunning{Y. Wu et al.}

\maketitle

\begin{abstract}
Unsupervised depth completion and estimation methods are trained by minimizing reconstruction error. Block artifacts from resampling, intensity saturation, and occlusions are amongst the many undesirable by-products of common data augmentation schemes that affect image reconstruction quality, and thus the training signal. Hence, typical augmentations on images viewed as essential to training pipelines in other vision tasks have seen limited use beyond small image intensity changes and flipping. The sparse depth modality in depth completion have seen even less use as intensity transformations alter the scale of the 3D scene, and geometric transformations may decimate the sparse points during resampling. We propose a method that unlocks a wide range of previously-infeasible geometric augmentations for unsupervised depth completion and estimation.
This is achieved by reversing, or ``undo''-ing, geometric transformations to the coordinates of the output depth, warping the depth map back to the original reference frame.
This enables computing the reconstruction losses using the original images and sparse depth maps, eliminating the pitfalls of naive loss computation on the augmented inputs and allowing us to scale up augmentations to boost performance.
We demonstrate our method on indoor (VOID) and outdoor (KITTI) datasets, where we consistently improve upon recent methods %
across both datasets as well as generalization to four other datasets. Code available at: \href{https://github.com/alexklwong/augundo}{https://github.com/alexklwong/augundo}.
\end{abstract}

\section{Introduction}

Data augmentation is essential to training machine learning models; it plays a role in performance and generalization \cite{perez2017effectiveness,shorten2019survey,taylor2018improving}.  One common axiom of choosing augmentations is that the task output should remain invariant to the augmentation. For example, image flipping is a viable augmentation for classifying animals, since it does not alter the label. Conversely, flipping road signs can alter their meanings; hence, such augmentation can be detrimental to tasks involving road sign recognition. 
For geometric tasks, the range of augmentations is more restricted due to constraints of problem formulation: Stereo assumes pairs of frontoparallel rectified images; hence, in-plane rotations are not viable. Image-guided sparse depth completion relies on sparse points to ground estimates to metric scale; therefore, intensity transformations on sparse depth maps that alter the scale of the 3-dimensional (3D) scene are infeasible. Unsupervised learning of depth completion and estimation further limit the use of augmentations as the supervision signal comes from reconstructing the inputs, where augmenting the input introduces artifacts that impact reconstruction quality and therefore the supervision (see Sec. \suppmatref{D}, Fig. \suppmatref{1}, Tab. \suppmatref{8}, \suppmatref{9} in the Supp. Mat. for examples of artifacts and extended discussion on their effect on learning). Moreover, video-based unsupervised training assumes rigid motion, so augmentations that introduce padding (e.g., translation, rotation) will yield constant or edge extended borders across images (i.e., no motion), preventing the model from properly learning depth and pose -- leaving few augmentations viable. While simulating nuisances is desirable, naively applying augmentations may do more harm than good; thus, it is unsurprising that existing work in unsupervised depth completion 
\cite{ma2019self,liu2022monitored,shivakumar2019dfusenet,wong2020unsupervised,wong2021learning,wong2021adaptive,wong2021unsupervised,yang2019dense} and estimation \cite{godard2017unsupervised,godard2019digging,lyu2021hr,zhang2023lite,zhou2017unsupervised} primarily rely on a small range of photometric augmentations and flipping.

Nevertheless, photometric augmentations help model the diverse range of illumination conditions and colors of objects that may populate the scene; geometric augmentations can simulate the various camera parameters, i.e., image resizing (zooming) can model changes in focal length, and scene arrangements, i.e., image flipping. However, block artifacts, loss during resampling, and intensity saturation are just some of the many undesirable side-effects of traditional augmentations to the image and sparse depth map for unsupervised learning of geometric tasks. To avoid  compromising the supervision signal, we %
compute the typical reconstruction loss on the original input image and sparse depth map instead of the augmented inputs, which bypasses negative effects of reconstruction artifacts due to photometric and geometric augmentations. However, there exists a mis-alignment between the original input (e.g., image, sparse depth), and the model depth estimate as geometric augmentations induce a change in coordinates. Hence, we \emph{undo} the geometric augmentations by inverting them in the output space to align the model estimate with the training target. 
 
Amongst the many geometric tasks, we focus on \textit{unsupervised depth completion}, the task of inferring a dense depth map from an image and sparse depth map, where augmentations have seen limited use. Here, a training sample includes the input sparse depth map, its associated image and additional images of neighboring views of the same 3D scene. Our method is also applicable to \textit{unsupervised monocular depth estimation}, which omits the sparse depth modality. Augmentations have traditionally been restricted to a limited range of photometric transformations and flipping -- due to the need to preserve photometric consistency across a sequence of video frames used during training, and the sparse set of 3D points projected onto the image frame as a 2.5D range map; degradation to either modalities directly impacts the supervision signal as the loss function is conventionally computed on the augmented inputs. By using our method, loss functions involving sparse depth and image reconstruction from other views can be computed on the original inputs while applying augmentations that were previously not viable for the task. 
\textbf{Our hypothesis:} By ``undo-ing'' the augmentations, one can expand the viable set and scale up their use in training, leading to improved model performance and generalizability. 

\begin{figure*}[t]
    \centering
    \includegraphics[width=0.73\textwidth]{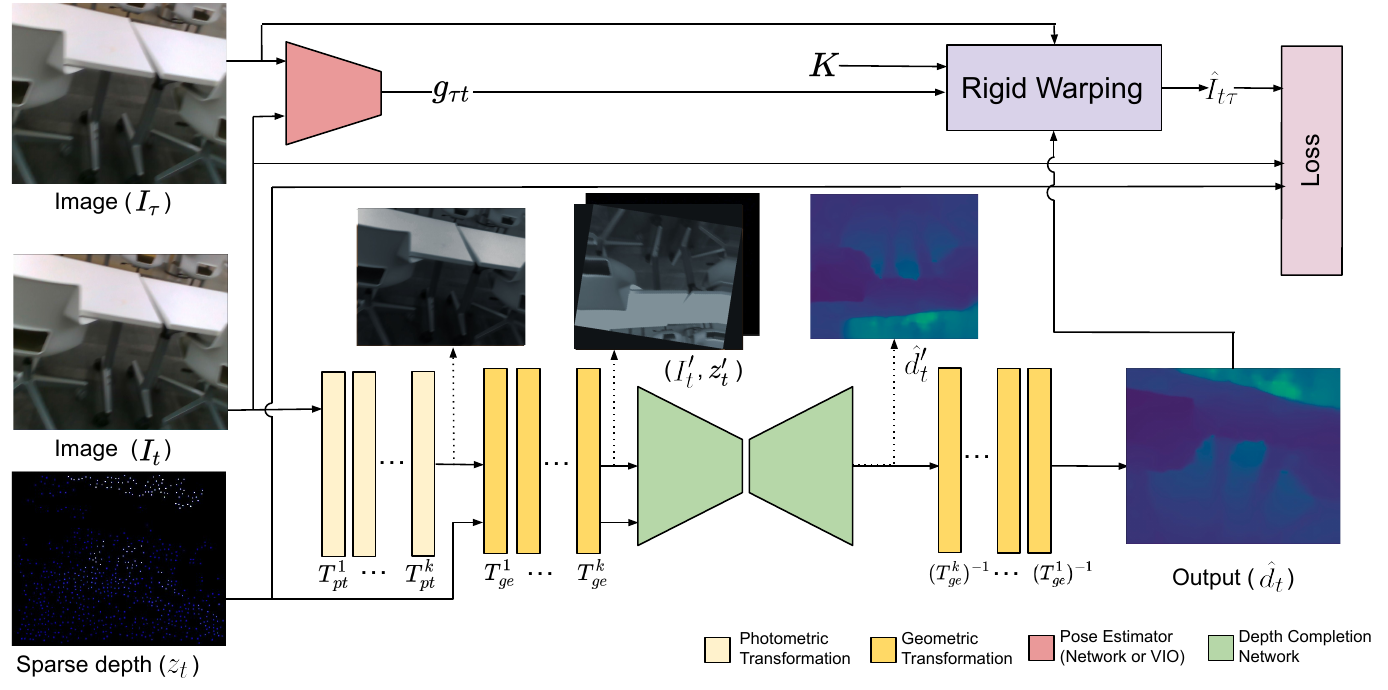}
    \caption{\textit{Overview.} We apply photometric augmentations to the input image, and the same set of geometric augmentations to both the input image and sparse depth map. We warp the output depth back to the original reference frame with the inverse geometric transformations. This enables image and sparse depth reconstruction losses to be computed on the original inputs while unlocking previously-infeasible augmentations. %
    }
    \label{fig:overview}
\end{figure*}

To this end, we introduce AugUndo, an augmentation framework that enables one to apply a wide range of photometric and geometric transformations on the inputs, and to ``undo'' them during loss computation. This allows computing the unsupervised loss on the original images and sparse depth maps, free of artifacts, through a warping of the output depth map -- obtained from augmented input -- onto the input frame of reference based on the inverse geometric transformation. In addition to group transformations that allow for output alignment, we combine them with commonly employed photometric augmentations. %
To the best of our knowledge, we are the first to propose a unified augmentation scheme for photometric and geometric augmentations for unsupervised depth completion and estimation.
We demonstrate AugUndo on recent methods on indoor and outdoor settings, where we consistently improve all methods across all datasets. %

\textbf{Our contributions} are as follows: (1) We propose AugUndo, a simple-yet-effective framework to scale up photometric and geometric augmentations for unsupervised depth completion and estimation, without compromising the supervision signal; AugUndo can be applied in a plug-and-play manner to existing methods with negligible increase in computational costs during training. 
(2) 
We enable previously-infeasible augmentations to be used for training unsupervised methods and comprehensively ablate combinations of eleven types of augmentations to study the performance benefits of each.
(3) We show that AugUndo can consistently improve robustness to shifts in sparse point densities for completion, model performance as well as zero-shot generalization for both depth completion and estimation for indoor and outdoor scenarios; thus, validating our hypothesis.

\section{Related Work}
\textit{Data augmentation for depth completion and estimation.}
While photometric transformations such as color jitter are often applied for unsupervised depth completion  \cite{ma2019self,wong2021learning,wong2021adaptive,wong2020unsupervised}, and estimation \cite{wong2019bilateral,fei2019geo, godard2017unsupervised,godard2019digging,guizilini20203d,poggi2020uncertainty,lao2024depth,lao2024sub,ranftl2020towards,zeng2024wordepth,zhan2018unsupervised},  geometric augmentations other than simple flipping are less commonly adopted. 
Most supervised depth completion methods \cite{eldesokey2020uncertainty,hu2021penet,jaritz2018sparse,li2020multi,lin2022dynamic,kam2022costdcnet,krishna2023deepsmooth,qiu2019deeplidar,qu2021bayesian,qu2020depth,zhang2023completionformer,yang2019dense,xu2019depth,zhang2018deep} similarly limit their augmentations to color jitter and horizontal flips due to sparse depth maps being decimated by rotation and scaling, causing points to be interpolated away. Nevertheless, for supervised training, it is straightforward to directly apply the same transformation to the ground truth for training. Indeed, some supervised depth completion methods \cite{cheng2018depth,cheng2020cspn++,park2020non,yu2023aggregating,van2019sparse,ezhov2024all,park2024test,singh2023depth} adopted random scaling, translation, in-plane rotation. Translation augmentations are also commonly applied in supervised depth estimation methods \cite{eigen2014depth,eigen2015predicting,laina2016deeper,li2015depth,liu2015learning,ranftl2020towards,ranftl2021vision,upadhyay2023enhancing,yang2024binding,yin2019enforcing,wong2020targeted,xu2017multi}. We posit that the reason that such transformations can be applied to the ground truth in supervised settings is due to artifacts caused by the transformation of a piece-wise smooth depth map being less severe than those of an RGB image and its intensities. However, such assumptions do not hold in unsupervised training. These artifacts would affect the training signal for unsupervised methods, which rely on photometric correspondences and observed sparse points. Our approach aims to resolve this to allow diverse geometric augmentations %
to be applied in a plug-and-play fashion in unsupervised training.

\textit{Unsupervised depth completion} methods \cite{ma2019self,shivakumar2019dfusenet,wong2021learning,wong2021adaptive,wong2020unsupervised,yan2023desnet,yang2019dense} leverage image and sparse depth reconstructions as training signals by minimizing errors between the input image and its reconstruction from other views, and errors between the input sparse depth map and the predicted depth along with a local smoothness regularizer. \cite{ma2019self} used Perspective-n-Point \cite{lepetit2009epnp} and RANSAC \cite{fischler1981random} to align consecutive video frames. \cite{yang2019dense} learns a depth prior conditioned on the image. \cite{wong2021learning} uses synthetic data to learn a prior on the shapes populating a scene, while \cite{lopez2020project} translates synthetic data to real domain to leverage rendered depth. \cite{wong2021adaptive} proposed an adaptive scheme to reduce penalties incurred on occluded regions. \cite{wong2021unsupervised} maps the image onto the 3D scene through calibrated back-projection. \cite{yan2023desnet} decouples structure and scale. \cite{jeon2022struct} uses line features from visual SLAM and \cite{liu2022monitored} introduced monitored distillation for positive congruent training. Augmentations for these methods are limited to a small range of photometric perturbations and image flipping. Operations such as rotation, resizing, and translation require resampling, which creates artifacts and affects the reconstruction quality. This causes performance degradation since loss is typically computed on the augmented images. Loss in sparse depth maps is further impacted as resampling and interpolation may cause loss of sparse points.  Contrary to these limited augmentation schemes, we enable a large range of photometric and geometric augmentations to be introduced during training. 

\textit{Unsupervised monocular depth estimation}, like depth completion, also minimizes photometric reconstruction error. \cite{garg2016unsupervised} frames depth estimation as a novel view synthesis problem. %
\cite{godard2017unsupervised} improves \cite{garg2016unsupervised} by imposing a consistency loss on the depth predicted from left and right images. \cite{zhou2017unsupervised} uses a pose network to enable unsupervised training on video sequences. \cite{godard2019digging} introduced auto-masking and min reprojection loss. To improve the supervision signal,  \cite{tosi2019learning,watson2019self,choi2021adaptive} leverage noisy proxy labels and \cite{poggi2018learning} uses a trinocular assumption. Additional loss terms based on visual odometry \cite{wang2018learning}, iterative closest point \cite{mahjourian2018unsupervised}, surface normals \cite{yang2018lego}, and semantic segmentation \cite{guizilini2019semantically,kumar2021syndistnet} were also introduced; \cite{poggi2020uncertainty,yang2020d3vo} further included predictive uncertainty. To handle rigid and non-rigid motions in the scene, previous works explored multi-task learning to include optical flow and moving object estimation \cite{chen2019self,ranjan2019competitive,yang2018every,yin2018geonet,zou2018df}, and used semantics to filter out outlier regions \cite{klingner2020self}.
\cite{lyu2021hr} redesigned the skip connection and decoders to extract high-resolution features, \cite{zhao2022monovit} combined global and local representations and \cite{zhang2023lite} introduced a lightweight architecture with dilated convolution and attention. In addition to depth completion, our method also shows consistent improvements on monocular depth estimation for \cite{godard2019digging,lyu2021hr,zhang2023lite}.

\section{Method Formulation}\label{sec:method}
Let $I : \Omega \subset \mathbb{R}^2 \rightarrow \mathbb{R}^3_+$ be an RGB image captured by a calibrated camera, $z : \Omega_z \subset \Omega \rightarrow \mathbb{R}_+$ the corresponding sparse point cloud projected onto the image plane as a sparse depth map, and $K \in \mathbb{R}^{3 \times 3}$ the intrinsic calibration matrix. %
Given an image and its sparse depth map, depth completion aims to learn a function $f_\theta(I, z)$ that recovers the distance between the camera to points in the 3D scene as a dense depth map. %
In another mode, if sparse depth maps are not given, then the problem reduces to monocular depth estimation which learns a function $f_\theta(I)$ to map a single image to a depth map $\Omega \rightarrow \mathbb{R}_+$. 
For the ease of notation, we denote the output depth map for both depth completion and estimation as $\hat{d} \in \mathbb{R}_+^{H \times W}$ where $H$ and $W$ are its height and width.

Unsupervised depth completion relies on photometric and sparse depth  reconstruction errors as its primary supervision signal. Without loss of generality, we assume an input of $(I_t, z_t)$ for an RGB image and associated sparse depth map captured at time $t$ and during training, an additional set of temporally adjacent images $I_{\tau}$ for $\tau \in \Upsilon \doteq \{t-1, t+1\}$. The reconstruction $\hat{I}_{t \tau}$ of $I_t$ from image $I_\tau$ is given by the reprojection based on estimated depth $\hat{d}_t := f_\theta(\cdot)$
\begin{equation}
    \hat{I}_{t \tau} (x) = I_\tau (\pi g_{\tau t} K^{-1} \bar{x} \hat{d}_t(x))
\label{eqn:image_reconstruction}
\end{equation}
where $\bar{x} = [x^\top, 1]^\top$ is the homogeneous coordinates of $x \in \Omega$,  $g_{\tau t} \in SE(3)$ the relative pose of the camera from time $t$ to $\tau$, $K$ the intrinsic calibration matrix, and $\pi$ the canonical perspective projection.

Using \eqnref{eqn:image_reconstruction}, a depth completion or estimation network $f_\theta$ minimizes
\begin{align}
    \argmin_\theta \sum_{\tau\in \Upsilon} \sum_{x \in \Omega} \alpha \rho\big(\hat{I}_{t \tau}(x), I_t(x)\big) + 
    \sum_{x \in \Omega_z}  \beta \psi\big(\hat{d}_t(x), z_t(x)\big) + \lambda R(\hat{d_t}) 
\label{eqn:loss_function}
\end{align}
where $\rho$ denotes the photometric reconstruction error, typically $L_1$ difference in pixel values and/or structural similarity (SSIM), $\psi$ the sparse depth reconstruction error, typically $L_1$ or $L_2$, $R$ the regularizer that biases the depth map to be piece-wise smooth with depth discontinuities aligned with edges in the image (commonly used by previous works \cite{godard2019digging,ma2019self,wong2020unsupervised,wong2021unsupervised}), and $\alpha$, $\beta$ and $\lambda$ their respective weighting. Note: monocular depth omits sparse depth term, i.e., $\beta = 0$.

\textit{Inverse transformation.} Let $\mathcal{A}_{pt}$ be the set of possible photometric transformations, and $\mathcal{A}_{ge}$ be the set of all geometric transformations. Given $\ T_{ge} \in \mathcal{A}_{ge}$, we wish to obtain an inverse transformation $T_{ge}^{-1}$ such that $T_{ge} \circ T_{ge}^{-1} \approx Id$ the identity function\footnote{Note that in practice, not all transformations are bijective. For instance, image rescaling with a fixed size image plane; hence strict equality is not always possible.}.
At each time step, we can sample a sequence of transformations $\{ T_{pt}^1 \ldots T_{pt}^k \}$ and $\{ T_{ge}^1 \ldots T_{ge}^m \}$ respectively from $\mathcal{A}_{pt}$ and $\mathcal{A}_{ge}$ to construct transformations $T_{pt} = T_{pt}^1 \circ \cdots \circ T_{pt}^k$ and $T_{ge} = T_{ge}^1 \circ \cdots \circ T_{ge}^m$.
We denote the composition of a collection of augmentation transformations by $T = T_{ge} \circ T_{pt} $ where $T_{pt} $ denotes photometric transformation, and $T_{ge}$ denotes geometric transformation. Furthermore, we denote the inverse geometric transformations by $T_{ge}^{-1} = ({T_{ge}^m})^{-1} \circ (T_{ge}^{m-1})^{-1} \circ \cdots \circ ({T_{ge}^1})^{-1}$, which operates on the space of depth maps to reverse the geometric transformation so that we can warp the output depth map onto the reference frame of the original image. Through inverse warping, this process is  differentiable.
In practice, some transformations cause border regions of the image to be warped out of frame, i.e., translated off the image plane or cropped away. Hence, after warping our output depth map back to the original frame of reference using the inverse geometric transformations, border extensions (edge paddings) are introduced to handle out-of-frame regions.

\textbf{AugUndo.} We apply each geometric transformation over  image coordinates and resample (bilinear for image and nearest neighbor for sparse depth):
\begin{align}
    \label{eqn:geometric_transform}
    \begin{bmatrix} x' & 1 \end{bmatrix}^\top &= T_{ge} \begin{bmatrix}x & 1\end{bmatrix}^{\top} \\
    I'(x') &= T_{pt}(I)(x); \ z'(x') = z(x)
    \label{eqn:resample_image_sparse_depth}
\end{align}
where $T_{ge}$ is the geometric transformation, $x \in \Omega$ and $x'
\in \Omega$ are coordinates in the image grid, and $I'$ is the image after the transformation; for ease of notation, we hereafter denote $I' = T(I) = T_{ge} \circ T_{pt} (I)$ to include both augmentations
through composition. Note that $x$ is in the original image reference frame and $x'$ is in the transformed image reference frame. Naturally, this process can be extended to multiple geometric transformations by composing them, i.e., $T_{ge} = T^{1}_{ge} \circ T^{2}_{ge} \circ \cdots \circ T^{m}_{ge}$. The reverse process is simply inverting the transformations where 
$T^{-1}_{ge} = (T^{m}_{ge})^{-1} \circ (T^{m-1}_{ge})^{-1} \circ \cdots \circ (T^{1}_{ge})^{-1}$:
\begin{align}
    \label{eqn:inverse_geometric_transform}
    \begin{bmatrix} x'' & 1\end{bmatrix}^\top &= T^{-1}_{ge} \begin{bmatrix}x' & 1 \end{bmatrix}^{\top} \\ 
    \hat{d}(x'') &= \hat{d}'(x')
\label{eqn:resample_output_depth}
\end{align}
where $\hat{d}'$ is the depth map inferred from augmented inputs $(I', z')$. Once reverted back to the original reference frame, $\hat{d}$ is used to reconstruct $I_t$ from $I_\tau$ for $\tau \in \Upsilon$ as in \eqnref{eqn:image_reconstruction}. More details can be found in Alg. \suppmatref{1} in the Supp. Mat.

By modeling $T^{-1}_{ge}$, \cref{eqn:geometric_transform}-\cref{eqn:resample_output_depth} allow us to apply a wide range of augmentations, while still establishing the correspondence between $I_t$ and $I_\tau$. Specifically, for minimizing  \cref{eqn:loss_function}, one may simply augment the image and sparse depth by $T$ to yield $(I'_t, z'_t)$ as input, and  reconstruct the original image and sparse depth $(I_t, z_t)$ from other views $I_\tau$ and the aligned output depth $\hat{d}$ (\cref{eqn:resample_output_depth}). We note that the inverse transformation is critical for enabling the sparse depth reconstruction term to be computed properly in \cref{eqn:loss_function}; if computed in the transformed reference frame, i.e., on $z'_t$, many of the sparse points would be decimated by interpolation during rotation and resizing (downsampling) augmentations -- leaving a lack of supervision from the sparse depth term of the loss function. We note that the original image sequence is fed to the pose estimator, rather than the augmented images. This ensures the estimated relative pose used during reprojection (\cref{eqn:image_reconstruction}) also corresponds to the original images.

\textit{Modeling geometric augmentations.} Depth completion differs from typical image-based (e.g., monocular) depth estimation, where operations like resizing are susceptible to scale ambiguity and can either correspond to changes in focal length, or distance from the camera (i.e., due to changes in camera pose). In depth completion, the sparse depth map grounds image pixels to specific depth values. Hence, there is no ambiguity in 
distance of each sparse point from the camera. Our goal is to synthesize new training data via geometric transformations that are consistent with the original 3D scene. Given that there is no scale ambiguity, this can be achieved by either (1) changes in camera pose (extrinsics), or (2) changes in camera parameters (intrinsics).

Augmentation via (1) would require warping the inputs for view synthesis. Since we operate under the unsupervised setting, accurate warping (preserving 3D scene) without access to dense ground truth depth values is difficult to achieve. An exception to this is rotation,  which we address below. On the other hand, (2) can be achieved via standard 2D image transformations while preserving the 3D structure of the scene. This motivates our modeling of geometric transformations, i.e., resizing, translation, as changes in camera parameters. 

For example, augmenting focal length (i.e. zooming in/out) is akin to ``resizing''. Similarly, translation can be used to model shifting of the optic center, i.e. capturing the same 3D scene using cameras with different principal point offsets. Note: image rotation can be seen as rotating a camera about its optical axis, which does not change the distance of a point from the camera. Thus, the camera position and the scene are kept constant across augmentations, eliminating the need for adjusting depth values via warping during the process. Instead, we only have to realign the output depth back to the original frame of reference (\cref{eqn:inverse_geometric_transform,eqn:resample_output_depth}), which makes our method computationally efficient. %

\textbf{Augmentations.} Our choices are based on common nuisance variabilities, i.e., changes in illumination, occlusions, and object color, scene layout (flip), and camera parameters (resize, translation) and orientation (rotation).

\textit{Photometric.} We include brightness, contrast, saturation, and hue, where all follow standard augmentation pipeline in existing works \cite{godard2019digging,wong2020unsupervised,wong2021unsupervised,zhang2023lite}. Applying the inverse of photometric augmentation can be viewed as recovering the original image; hence, we directly use original image instead of applying transformations to the intensities (which are not recoverable if saturated at the max value).

\textit{Occlusion.} We consider patch removal and sparse point removal. For the former, we randomly select a percentage of pixels $x \in \Omega$ in the image and remove (with zero-fill) an arbitrary-sized patch around it. For the latter, we randomly sample a percentage of points $x \in \Omega_z$ in the sparse depth map. The inverse transformation of this is simply the original image and sparse depth map, both used in loss computation before augmentation.

\textit{Flip.} We consider horizontal and vertical flips. When applied, the same flip operation is used for both input image and sparse depth maps. We record the flip type during data augmentation. During loss computation, we reverse the flip direction to align output depth with the original image and sparse depth map.

\textit{Resize.} We define a new image plane of the input resolution and generate a random scaling factor to be applied along both height and width directions. The image is warped to the new image plane, where any point warped out of the plane is excluded; borders of the warped image are extended to the bounds of the image plane by edge replication. 
Sparse points occupying multiple pixels are eroded to a single point for resizing with a factor greater than 1.
During loss computation, we warp the output depth map onto a new image plane of the same dimensions by the inverse scaling matrix of the recorded scaling factors; borders of the warped depth map are extended by edge replication if necessary.
We view resizing factors greater and less than 1 (zooming in and out) as two separate forms of augmentations to distinguish their contributions.

\textit{Rotation.} Naive rotation leads to the loss of large areas of the image, i.e., cropped away to retain the same-size image, discarding large portions of possible co-visible regions and also supervisory signals. To preserve the entire image, we first warp the image by a randomly generated angle to a new (larger) image plane, so that the rotated image fits tightly within the new image. As image sizes within a batch can vary depending on the rotation angle, we center-pad (uniformly on each side) each image in the batch to the maximum width and height of the augmented batch.  To reverse the rotation on the output depth, we warp the output depth map back by the inverse rotation matrix, then perform a center crop on the depth map to align with the original input.

\textit{Translation.} We define a new image plane of the same dimensions as the input and generate random height and width translation factors. The coordinates of the input are translated and its pixels are inverse warped onto the new image plane. Any pixel warped out of the image plane is excluded. Borders of the warped image are extended to the bounds of the image plane by edge replication. During loss computation, we warp the output depth map onto a new image plane of the same dimensions by the inverse translation matrix. Borders of the warped depth map are extended to the bounds of the image plane by edge replication. %

\section{Experiments}
\label{sec:experiments}
We demonstrate AugUndo on recent unsupervised depth completion (VOICED \cite{wong2020unsupervised}, FusionNet \cite{wong2021learning}, KBNet \cite{wong2021unsupervised}) and estimation (Monodepth2 \cite{godard2019digging}, HRDepth \cite{lyu2021hr}, LiteMono \cite{zhang2023lite}) methods on two  datasets (KITTI \cite{geiger2012we, uhrig2017sparsity}, VOID \cite{wong2020unsupervised}) here. \textit{Due to space limitations}, we defer the results of MonDi \cite{liu2022monitored} and DesNet \cite{yan2023desnet} to Sec. \suppmatref{K} in the Supp. Mat. %
To evaluate zero-shot generalization, we test on three additional datasets for each task: NYUv2 \cite{silberman2012indoor}, ScanNet \cite{dai2017scannet}, Waymo \cite{sun2020scalability}, and Make3d \cite{saxena2008make3d} -- where we transfer models on VOID to NYUv2 (\tabref{tab:depth_completion_zero_shot}) and ScanNet, and from KITTI to Waymo (Tab. \suppmatref{4} in Supp. Mat.) for completion and Make3d \cite{saxena2008make3d} (Tab. \suppmatref{5} in Supp. Mat.) for estimation. We also test model sensitivity to different sparse depth input densities in \tabref{tab:depth_completion_void} (right) -- see Sec. \suppmatref{H} of the Supp. Mat for an extensive study. Additionally, we present an comprehensive ablation study in Sec. \suppmatref{G} in the Supp. Mat. to test the effect of each augmentation. 
We also provide results on modeling AugUndo as changes in camera motion or parameters in Tab. \suppmatref{7} of the Supp. Mat.; interpreting geometric augmentations as different camera parameters yields better performance. Tabs. \suppmatref{8}, \suppmatref{9} in Supp. Mat. show that unsupervised models cannot be trained with naive geometric augmentations. See Sec. \suppmatref{E} of the Supp. Mat. for descriptions of datasets.

\textit{Experiment setup.} We followed the original settings of the open-sourced code for each work and modified the data handling and loss function to incorporate AugUndo. We perform 4 independent trials for each experiment and report their means and standard deviations. To ensure a fair comparison, we train all models from scratch. %
Below, we report the best combination of augmentations found empirically through extensive experiments on each dataset. See Secs. \suppmatref{A} and \suppmatref{B} in Supp. Mat. for implementation details, and evaluation metrics, respectively.

\textit{Augmentations.} Through a search over augmentation types and values, we found a consistent set that tends to yield improvements across all methods with small changes to degree of augmentation catered to each method. See Sec. \suppmatref{C} in Supp. Mat. for details of augmentation parameters. Rows in \cref{tab:depth_completion_void,tab:depth_estimation_void,tab:depth_completion_zero_shot,tab:monocular_depth_prediction_generalization_indoors,tab:depth_completion_kitti,tab:depth_estimation_kitti} marked with "+ AugUndo" denote models trained with our method.  We note that performance gain can be obtained by typical set and ranges of augmentations and does not require a meticulous selection of hyper-parameters (see Sec. \suppmatref{J} of Supp. Mat. for a sensitivity study). Results reported here aims to study how far one can push performance and generalization, purely from augmentations.

\begin{figure*}[t]
    \centering
    \includegraphics[width=0.95\textwidth]{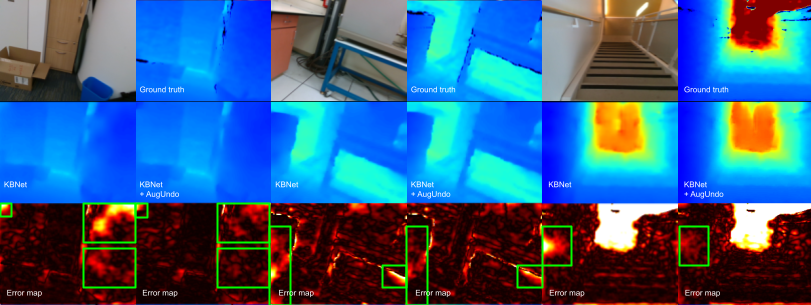}
    \caption{\textit{Depth completion on VOID}. We compare KBNet trained with standard augmentations and with AugUndo. AugUndo consistently produces lower errors with reduced border artifacts and improves on homogeneous regions, i.e., pillar (left), cabinet (middle), wall of staircase (right). Error maps are highlighted for comparison. %
    }
    \label{fig:qualitative_result_void_kbnet}
\end{figure*}

\textbf{Results on VOID.} We begin by presenting quantitative results for AugUndo on unsupervised depth completion. \tabref{tab:depth_completion_void} (left, VOID1500) shows our main results on the VOID benchmark. By training the models with AugUndo, we observe an average overall improvement of 18.3\% across all methods and metrics on VOID1500.
Specifically, we improve VOICED by 26.4\%, FusionNet by 16.3\%, and KBNet by 12.2\%. These experiments validate our hypothesis that by applying a wider range of augmentations, we are able to improve the baseline model's performance. They also illustrate the lack in use of augmentations in existing works: incorporating standard augmentations (albeit requiring modification to augmentation and loss computation pipelines) can yield a large performance boost. \figref{fig:qualitative_result_void_kbnet} shows a head-to-head comparison between KBNet trained using standard procedure in \cite{wong2021unsupervised} and KBNet trained using AugUndo. We observe qualitative improvements from AugUndo where we improve KBNet in homogeneous regions and image borders, i.e., pillar (left), cabinet (middle), wall (right). Applying geometric augmentations yields models with fewer border artifacts as we apply random translation, which warps part of the image out of frame. This allows us to simulate border occlusion, where there lacks correspondence in an adjacent frame. While training with standard protocol results in failures to recover structures near the image border, training with AugUndo can render models robust to them by computing the loss on the original frame of reference (where we do have correspondence) through our inverse transformations of the depth map. Additionally, resizing allows the model to learn multiple resolutions of the input, akin to zooming in and out, which can also impose smoothness in homogeneous regions through the scale-space transition; whereas, rotation can simulate diverse camera orientations. \figref{fig:qualitative_result_void_kbnet} shows that translation, rotation, and resizing can model these effect in the input space to yield models robust to these nuisance variability. 

\begin{table}[t]
  \centering
      \caption{
        \textit{Depth completion on VOID.} %
        AugUndo improves performance by an average of 18.3\% across all methods and metrics on VOID1500. When models trained on VOID1500 are tested VOID500, AugUndo improves by 23.1\%, as translation, resizing, and occlusion augmentations vary sparsity by removing sparse points from the input.
    }
    \scalebox{0.73}{
    \begin{tabular}{l c c c c c c c c}
        \midrule
         & \multicolumn{4}{c}{VOID1500} & \multicolumn{4}{c}{VOID500} \\
         \cmidrule(lr){2-5} \cmidrule(lr){6-9}
         Method & MAE $\downarrow$& RMSE $\downarrow$& iMAE $\downarrow$& iRMSE $\downarrow$ & MAE $\downarrow$& RMSE $\downarrow$& iMAE $\downarrow$& iRMSE $\downarrow$ \\ 
        \midrule
        VOICED
        & 74.78$\pm$2.69 & 139.75$\pm$4.57 & 39.20$\pm$1.46 & 71.98$\pm$2.54 
        & 137.01$\pm$4.23 & 235.80$\pm$7.82 & 71.36$\pm$1.86 & 130.63$\pm$5.66 \\
        + AugUndo					
        & \textbf{52.73}$\pm$0.41 & \textbf{111.09}$\pm$0.92 & \textbf{26.93}$\pm$0.54 & \textbf{54.46}$\pm$0.38
        & \textbf{92.99}$\pm$1.11 & \textbf{176.94}$\pm$1.38 & \textbf{46.43}$\pm$0.85 & \textbf{91.10}$\pm$1.64 \\ 
        \midrule
        FusionNet
        & 52.11$\pm$0.44 & 113.30$\pm$1.18 & 28.53$\pm$0.52 & 58.79$\pm$2.01
        & 97.73$\pm$0.73 & 194.32$\pm$1.36 & 58.65$\pm$1.31 & 122.95$\pm$3.04 \\
        + AugUndo
        & \textbf{41.16}$\pm$0.18 & \textbf{99.21}$\pm$0.39 & \textbf{22.23}$\pm$0.35 & \textbf{53.07}$\pm$1.30
        & \textbf{74.97}$\pm$0.69 & \textbf{162.71}$\pm$2.39 & \textbf{40.44}$\pm$1.39 & \textbf{92.11}$\pm$5.79 \\
        \midrule
        KBNet        
        & 38.11$\pm$0.77 & 95.22$\pm$1.72 & 19.51$\pm$0.14 & 46.70$\pm$0.48 & 78.44$\pm$1.39 & 178.17$\pm$3.27 & 37.56$\pm$0.61 & 83.43$\pm$1.89 \\
        + AugUndo
        & \textbf{33.32}$\pm$0.18 & \textbf{85.67}$\pm$0.39 & \textbf{16.61}$\pm$0.29 & \textbf{41.24}$\pm$0.60 		
        &\textbf{66.97}$\pm$0.81 & \textbf{151.55}$\pm$2.03 & \textbf{31.63}$\pm$0.53 & \textbf{71.90}$\pm$0.82 \\
        \midrule
    \end{tabular}}
\label{tab:depth_completion_void}
\end{table}

AugUndo also is applicable for monocular depth estimation. \tabref{tab:depth_estimation_void} shows a comparison using the standard augmentation procedure of Monodepth2, HR-Depth, and Lite-Mono and using AugUndo. \tabref{tab:depth_estimation_void} shows that AugUndo consistently improves all models across all error and accuracy metrics. Thus, validating the hypothesis that AugUndo can be applied generically to improve monocular depth estimation. Specifically, we observe a boost in the most difficult accuracy metric ($\delta < 1.25$), where Monodepth2 improves from 0.717 to 0.724, HR-Depth from 0.714 to 0.718 and Lite-Mono from 0.669 to 0.674. At the same time, for AbsRel metric, Monodepth2 improves from 0.183 to 0.178, HR-depth improves from 0.185 to 0.181, and Lite-Mono improves from 0.209 to 0.200. 

\begin{table}[t]
    \centering
        \caption{\textit{Monocular depth estimation on VOID}. AugUndo is applicable for monocular depth estimation and consistently improves three monocular depth models.}
    \setlength\tabcolsep{1pt}
    \scalebox{0.75}{\begin{tabular}{l c c c c c c c}
        \midrule
        Method & MAE $\downarrow$ & RMSE $\downarrow$& AbsRel $\downarrow$& SqRel $\downarrow$& $\delta < 1.25 \uparrow$ & $\delta < 1.25^2\uparrow$ & $\delta < 1.25^3\uparrow$ \\ 
        \midrule
        Monodepth2 
        & 283.861$\pm$3.732 & 395.947$\pm$5.728 & 0.183$\pm$0.002 & 0.100$\pm$0.003 & 0.717$\pm$0.005& 0.922$\pm$0.004 & 0.975$\pm$0.002 \\
         + AugUndo
        & \textbf{277.696}$\pm$4.861  & \textbf{388.088}$\pm$5.768& \textbf{0.178}$\pm$0.003& \textbf{0.095}$\pm$0.004 & \textbf{0.724}$\pm$0.007 & \textbf{0.925}$\pm$0.004 & \textbf{0.978}$\pm$0.002 \\
        \midrule
        HR-Depth 
        & 286.282$\pm$7.059 & 399.112$\pm$9.184 & 0.185$\pm$0.004 & 0.100$\pm$0.004 & 0.714$\pm$0.012 & 0.919$\pm$0.006& 0.975$\pm$0.002 \\
        + AugUndo
        & \textbf{283.086}$\pm$6.787 & \textbf{394.261}$\pm$9.133 & \textbf{0.181}$\pm$0.005 & \textbf{0.097}$\pm$0.005 & \textbf{0.718}$\pm$0.013 & \textbf{0.922}$\pm$0.004 & \textbf{0.977}$\pm$0.002 \\
        \midrule
        Lite-Mono		
        & 319.910$\pm$15.00& 446.005$\pm$22.97& 0.209$\pm$0.013& 0.129$\pm$0.019& 0.669$\pm$0.016& 0.892$\pm$0.011 & 0.963$\pm$0.006 \\
        + AugUndo				
        & \textbf{308.010}$\pm$0.859 & \textbf{426.626}$\pm$0.484 & \textbf{0.200}$\pm$0.003 & \textbf{0.114}$\pm$0.001 & \textbf{0.674}$\pm$0.005 & \textbf{0.901}$\pm$0.002 & \textbf{0.969}$\pm$0.002 \\
        \midrule
    \end{tabular}}
\label{tab:depth_estimation_void}
\end{table}

\textit{Sensitivity study.} One limitation of existing depth completion training pipelines is that there are little to no augmentations applied to sparse depth modality. However, in real-world applications, sparse depth has, in fact, high variability, i.e., features tracked in SLAM/VIO systems will vary depending on the scene (points are dropped or added to the state), and point cloud densities returned by a sensor will differ based on specifications. To further examine the effect of AugUndo on sparse depth, we study the sensitivity to changes in sparse depth density by testing models on VOID1500 ($\approx$1500 points) on VOID500 ($\approx$500). For the 3$\times$ reduction in sparse points, AugUndo improves robustness by an average of 23.1\% across all methods. %
These improvements result from the geometric and occlusion augmentations made possible via AugUndo, which greatly increases sparse depth variations, i.e., decimating them through resizing, re-orienting their configuration through rotation, translating them out of frame, and randomly occluding them, to avoid overfitting particular sparse depth configurations. 
We further push their limits in Sec. \suppmatref{H} of the Supp. Mat., where we test them on VOID150 with a 10$\times$ reduction in sparse points from the training set (VOID1500) and observe the same trend of improvements. This shows that AugUndo significantly improves robustness of models to changes in sparse depth.

\textit{Zero-shot generalization.} We test depth completion models trained on VOID on NYUv2 and ScanNet. \tabref{tab:depth_completion_zero_shot} shows an average of 23.2\% improvement on NYUv2 and 27.6\% on ScanNet. Applying AugUndo to VOICED greatly improve its generalizability to both NYUv2 and ScanNet. This is likely due to the scaffolding densification employed by VOICED, where the network can overfit to scaffoldings of particular sparse depth configurations and therefore, does not generalize well when presented with sparse depth maps with different configurations. Like our sensitivity study (\figref{fig:qualitative_result_void_kbnet}, VOID500), occlusion and geometric augmentations introduce variation into the sparse depth configurations and densities, which alleviates overfitting to specific point clouds or 3D scenes; hence, improving VOICED from 49.9\% and 52.6\% on NYUv2 and ScanNet, respectively. For FusionNet and KBNet, we still see nontrivial improvement: FusionNet improves by 8\% and 12.1\% on NYUv2 and ScanNet, and KBNet by 11.7\% and 18.3\%. This further validates our hypothesis that by applying a more diverse set of transformations, we are able to improve generalization to new datasets.

\begin{table}[t]
  \centering
      \caption{
        \textit{Zero-shot transfer from VOID to NYUv2 and ScanNet for depth completion.} %
        AugUndo improves generalization of models trained on VOID to novel datasets by an average of 25.4\% for all evaluation metrics (in millimeters) across both datasets.
    }
    \scalebox{0.7}{
    \begin{tabular}{l c c c c c c c c}
        \midrule
        & \multicolumn{4}{c}{NYUv2} & \multicolumn{4}{c}{ScanNet} \\
         \cmidrule(lr){2-5} \cmidrule(lr){6-9}
        Method & MAE $\downarrow$ & RMSE $\downarrow$ & iMAE $\downarrow$ & iRMSE $\downarrow$ & MAE $\downarrow$ & RMSE $\downarrow$ & iMAE $\downarrow$ & iRMSE $\downarrow$ \\ 
        \midrule
        VOICED 
        & 2240$\pm$143.90 & 2427$\pm$143.49 & 211$\pm$9.60 & 238$\pm$10.89 & 1562$\pm$136.79 & 1764$\pm$146.39 & 270$\pm$17.25 & 311$\pm$17.36 \\
        + AugUndo 
        & \textbf{990}$\pm$82.48 & \textbf{1181}$\pm$65.55 & \textbf{110}$\pm$6.11 & \textbf{132}$\pm$6.01 & \textbf{638}$\pm$44.74 & \textbf{791}$\pm$79.48 & \textbf{131}$\pm$7.68 & \textbf{170}$\pm$6.32 \\
        FusionNet
        & 132.24$\pm$2.12 & 236.16$\pm$4.59 & 28.68$\pm$0.42 & 61.87$\pm$1.20 & 109.47$\pm$3.01 & 206.33$\pm$6.11 & 55.45$\pm$1.56 & 122.52$\pm$2.04 \\	
        + AugUndo 
        & \textbf{124.93}$\pm$3.05 & \textbf{227.23}$\pm$4.96 & \textbf{25.70}$\pm$0.41 & \textbf{54.09}$\pm$0.87 & \textbf{100.64}$\pm$2.31 & \textbf{195.85}$\pm$5.85 & \textbf{45.98}$\pm$0.78 & \textbf{99.90}$\pm$5.00 \\
        KBNet 
        & 138.31$\pm$5.74 & 257.99$\pm$10.36 & 25.48$\pm$0.63 & 51.77$\pm$0.99 & 103.05$\pm$4.99 & 217.12$\pm$13.35 & 36.23$\pm$1.12 & 76.55$\pm$2.90 \\
        + AugUndo 
        & \textbf{118.60}$\pm$3.44 & \textbf{231.13}$\pm$8.85 & \textbf{22.06}$\pm$0.31 & \textbf{47.07}$\pm$0.70 & \textbf{82.53}$\pm$4.33 & \textbf{175.30}$\pm$11.13 & \textbf{29.87}$\pm$1.06 & \textbf{63.78}$\pm$1.30 \\
        \midrule
    \end{tabular}}
\label{tab:depth_completion_zero_shot}
\end{table}

\begin{table*}[t]
    \centering
        \caption{\textit{Zero-shot transfer from VOID to NYUv2 and ScanNet for depth estimation.} AugUndo consistently improves generalization (in meters) for monocular depth models.}
    \setlength\tabcolsep{1pt}
    \scalebox{0.7}{\begin{tabular}{l l c c c c c c c c c c c}
        \midrule 
        Dataset & Method & MAE $\downarrow$ & RMSE $\downarrow$& AbsRel $\downarrow$& SqRel $\downarrow$ & $\delta < 1.25 \uparrow$ & $\delta < 1.25^2\uparrow$ & $\delta < 1.25^3\uparrow$ \\ 
        \midrule
        \multirow{6}*{NYUv2} & Monodepth2 & 0.432$\pm$0.003 & 0.556$\pm$0.005 & 0.205$\pm$0.001 & 0.159$\pm$0.001 & 0.683$\pm$0.005 & 0.907$\pm$0.001 & 0.975$\pm$0.001 \\ 						
        & + AugUndo & \textbf{0.415}$\pm$0.005 & \textbf{0.537}$\pm$0.006 & \textbf{0.196}$\pm$0.002 & \textbf{0.148}$\pm$0.003 & \textbf{0.700}$\pm$0.008 & \textbf{0.915}$\pm$0.002 & \textbf{0.977}$\pm$0.001 \\ 
				
        & HR-Depth & 0.424$\pm$0.003 & 0.549$\pm$0.003 & 0.201$\pm$0.002 & 0.154$\pm$0.002 & 0.692$\pm$0.003 & 0.910$\pm$0.002 & \textbf{0.976}$\pm$0.001\\					
        & + AugUndo & \textbf{0.421}$\pm$0.009 & \textbf{0.542}$\pm$0.011 & \textbf{0.199}$\pm$0.005 & \textbf{0.152}$\pm$0.006 & \textbf{0.696}$\pm$0.009 & \textbf{0.913}$\pm$0.004 & \textbf{0.976}$\pm$0.001 \\    
        & Lite-Mono & 0.480$\pm$0.012 & 0.616$\pm$0.018 & 0.231$\pm$0.008 & 0.199$\pm$0.015 & 0.637$\pm$0.009 & 0.879$\pm$0.009 & 0.964$\pm$0.004 \\
         & + AugUndo & \textbf{0.468}$\pm$0.003 & \textbf{0.595}$\pm$0.003 & \textbf{0.225}$\pm$0.004 & \textbf{0.187}$\pm$0.005 & \textbf{0.646}$\pm$0.002 & \textbf{0.889}$\pm$0.003 & \textbf{0.968}$\pm$0.001 \\ 
        \midrule	
        \multirow{6}*{ScanNet} & Monodepth2 & 0.284$\pm$0.004& 0.368$\pm$0.005& 0.177$\pm$0.002& 0.097$\pm$0.002& 0.741$\pm$0.006& 0.931$\pm$0.003& 0.980$\pm$0.001 \\
         & + AugUndo & \textbf{0.270}$\pm$0.003& \textbf{0.351}$\pm$0.004& \textbf{0.169}$\pm$0.001& \textbf{0.088}$\pm$0.002& \textbf{0.759}$\pm$0.004& \textbf{0.937}$\pm$0.001& \textbf{0.982}$\pm$0.001 \\
        & HR-Depth & 0.282$\pm$0.004 & 0.366$\pm$0.005& 0.175$\pm$0.002& 0.095$\pm$0.002& 0.743$\pm$0.005& 0.929$\pm$0.003& 0.979$\pm$0.002 \\						
        & + AugUndo & \textbf{0.274}$\pm$0.007& \textbf{0.357}$\pm$0.009& \textbf{0.172}$\pm$0.005& \textbf{0.092}$\pm$0.005&  \textbf{0.754}$\pm$0.009& \textbf{0.935}$\pm$0.003& \textbf{0.981}$\pm$0.001 \\	
        & Lite-Mono & \textbf{0.296}$\pm$0.002& 0.388$\pm$0.007 & \textbf{0.185}$\pm$0.003 & 0.109$\pm$0.006 & \textbf{0.731}$\pm$0.002 & 0.921$\pm$0.002 & 0.976$\pm$0.001\\	
        & + AugUndo & \textbf{0.296}$\pm$0.001	& \textbf{0.382}$\pm$0.001 & \textbf{0.185}$\pm$0.001 & \textbf{0.105}$\pm$0.002 & 0.728$\pm$0.002 & \textbf{0.924}$\pm$0.001 & \textbf{0.977}$\pm$0.001 \\ 
        \midrule		
    \end{tabular}}
\label{tab:monocular_depth_prediction_generalization_indoors}
\end{table*}

For depth estimation, we tested Monodepth2, HR-Depth, Lite-Mono (trained on VOID) for zero-shot generalizability to NYUv2 and ScanNet. The results are shown in \tabref{tab:monocular_depth_prediction_generalization_indoors}, where training with AugUndo consistently improves generalization errors across all evaluation metrics for both NYUv2 and ScanNet. Notably, for RMSE metric, Monodepth2 improves from 0.556 to 0.537, HR-Depth from 0.549 to 0.542, Lite-Mono from 0.616 to 0.595. For ScanNet, similar improvement on RMSE can also be observed, where Monodepth2 improves from 0.368 to 0.351, HR-Depth from 0.366 to 0.357, and Lite-Mono from 0.388 to 0.382. The improvement in RMSE metric, which is sensitive to outliers, highlights AugUndo's ability to model different input data distributions, collected by a different camera of different orientation with different object colors and layouts.

\begin{table}[t]
    \centering
      \caption{
        \textit{Depth completion on KITTI.}  AugUndo consistently improves all methods for all metrics (in millimeters) by approximately 5.2\% on average.
    }
    \scalebox{0.7}{
    \begin{tabular}{l c c c c}
        \midrule
        Method & MAE $\downarrow$& RMSE $\downarrow$& iMAE $\downarrow$& iRMSE $\downarrow$ \\ 
        \midrule
        VOICED		
        & 318.59$\pm$7.74 & 1,213.60$\pm$17.49 & 1.30$\pm$0.05 & 3.72$\pm$0.04 \\
        + AugUndo	
        & \textbf{295.41}$\pm$0.30 & \textbf{1,159.27}$\pm$5.44 & \textbf{1.20}$\pm$0.02 & \textbf{3.49}$\pm$0.03 \\         
        \midrule
        FusionNet     			
        & 285.55$\pm$2.16 & 1,174.47$\pm$10.67 & 1.20$\pm$0.03 & 3.45$\pm$0.08 \\
        + AugUndo
        & \textbf{267.69}$\pm$1.85 & \textbf{1,157.07}$\pm$4.61 & \textbf{1.08}$\pm$0.02 & \textbf{3.19}$\pm$0.03 \\
        \midrule
        KBNet    
        & 263.90$\pm$3.63 & 1,130.66$\pm$6.22 & 1.05$\pm$0.01 & 3.24$\pm$0.04 \\ 
        + AugUndo
        & \textbf{256.37}$\pm$1.00 & \textbf{1,114.53}$\pm$3.79 & \textbf{1.01}$\pm$0.01 & \textbf{3.13}$\pm$0.03 \\
        \midrule
    \end{tabular}
    }
\label{tab:depth_completion_kitti}
\end{table}

\begin{figure*}[t]
    \centering
    \includegraphics[width=0.9\textwidth]{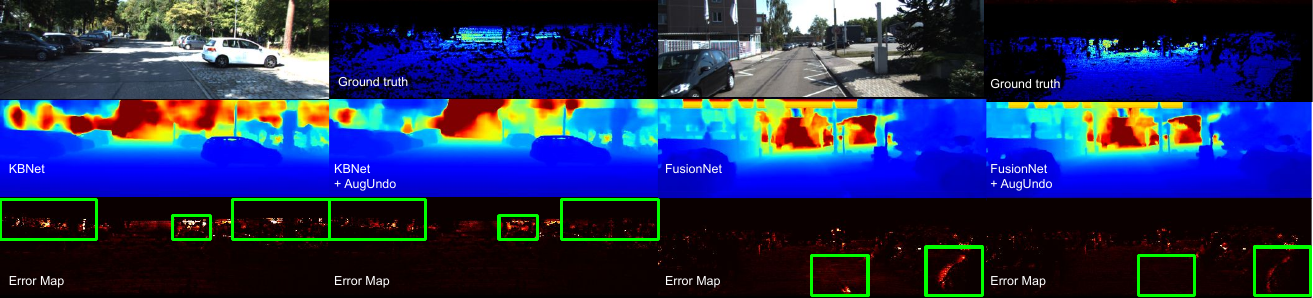}
    \caption{\textit{Depth completion on KITTI}. We compare KBNet and FusionNet trained with standard augmentations and with AugUndo. AugUndo consistently produces lower errors in highlighted regions where structures may have arbitrary orientation (vegetation) and regions near image border that typically lacks correspondence during training.}
    \label{fig:qualitative_result_kitti_kbnet_fusionnet}
\end{figure*}

\begin{table*}[t]
    \centering
        \caption{\textit{Monocular depth estimation on KITTI}. AugUndo consistently improve on evaluation metrics (in meters) across different models.}
    \setlength\tabcolsep{2pt}
    \scalebox{0.7}{\begin{tabular}{l c c c c c c c}
        \midrule
        Method & MAE $\downarrow$ & RMSE $\downarrow$& AbsRel $\downarrow$& SqRel $\downarrow$& $\delta < 1.25 \uparrow$ & $\delta < 1.25^2\uparrow$ & $\delta < 1.25^3\uparrow$ \\ 
        \midrule
        Monodepth2 &
        2.315$\pm$0.005 & 4.794$\pm$0.035 & 0.117$\pm$0.001 & \textbf{0.845}$\pm$0.030 & 0.869$\pm$0.004 & 0.959$\pm$0.001 & \textbf{0.982}$\pm$0.001  \\
        + AugUndo & 
        \textbf{2.237}$\pm$0.014 & \textbf{4.739}$\pm$0.032& \textbf{0.113}$\pm$0.000 & 0.862$\pm$0.030 &	\textbf{0.879}$\pm$0.002 & \textbf{0.960}$\pm$0.001 & \textbf{0.982}$\pm$0.001 \\
        \midrule 
        HR-Depth &
        2.226$\pm$0.004 & 4.626$\pm$0.032 & 0.113$\pm$0.001 & 0.797$\pm$0.022 & 0.879$\pm$0.002 & 0.961$\pm$0.001 & 0.982$\pm$0.000 \\
        +AugUndo &
        \textbf{2.185}$\pm$0.013 & \textbf{4.610}$\pm$0.029 & \textbf{0.111}$\pm$0.001 & \textbf{0.794}$\pm$0.021 & \textbf{0.883}$\pm$0.001 & \textbf{0.962}$\pm$0.001 & \textbf{0.983}$\pm$0.001 \\
        \midrule
        Lite-Mono &
        2.338$\pm$0.005 & 4.821$\pm$0.027 & 0.121$\pm$0.001 & 0.875$\pm$0.007 & 0.862$\pm$0.002 & \textbf{0.955}$\pm$0.001 & 0.980$\pm$0.000 \\
        + AugUndo &
        \textbf{2.314}$\pm$0.030 & \textbf{4.780}$\pm$0.055 & \textbf{0.120}$\pm$0.001 & \textbf{0.849}$\pm$0.019 & \textbf{0.863}$\pm$0.004 & \textbf{0.955}$\pm$0.001 & \textbf{0.981}$\pm$0.001 \\
        \midrule
    \end{tabular}}
\label{tab:depth_estimation_kitti}
\end{table*}

\textbf{Results on KITTI.}
We begin with quantitative results for depth completion. While AugUndo consistently improves all methods (\tabref{tab:depth_completion_kitti}), we note that the improvement is less, but respectable, in this case: $\approx$5.2\% overall with the largest gain in FusionNet of 6.41\%. This is largely due to the small scene variations in the outdoor driving scenarios, i.e., ground plane with vehicles, buildings on the sides, horizontal lidar scans, and largely planar motion. The dataset bias is strong enough to render vertical flip detrimental to performance. This is also evident in existing works as none utilizes vertical flip as augmentation. Nonetheless, AugUndo still improves performance, where point removal models different lidar returns patterns, and resizing simulates large variations in scales of objects observed in road scenes. Similar to indoors, translations help model occlusions by shifting the projection with different principal points. 
\figref{fig:qualitative_result_kitti_kbnet_fusionnet} compares FusionNet and KBNet trained using conventional augmentations and AugUndo. AugUndo improves over objects that may have diverse orientations (i.e., tree branches and vegetation), thanks to rotation augmentations. Improvements are also observed in ``small'' objects at far regions where resizing can simulate zooming in/out to emulate objects of different sizes. Through translation and occlusion, AugUndo also improves on occlusion boundaries during training (highlighted), making estimates near image borders and object boundaries more robust.

We additionally show results for monocular depth estimation on KITTI in \tabref{tab:depth_estimation_kitti}, where we compare the standard augmentation pipelines used by \cite{godard2019digging,lyu2021hr,zhang2023lite} and AugUndo. We observe similar trends in performance gain as we did in depth completion trained on KITTI: Applying our set of augmentations improves most metrics for all methods. We observe notable improvements in $\delta < 1.25$, the most difficult accuracy metric, from 0.869 to 0.879 in Monodepth2, from 0.879 to 0.883 in HR-Depth, and 0.862 to 0.863 in Lite-Mono. For Monodepth2, we also observe a large reduction in AbsRel error, improving it from 0.117 to 0.113. Similarly, for Lite-Mono, we also boosted performance across all metrics, with a particularly high reduction in SqRel from 0.875 to 0.849.

\textit{Remarks.} As AugUndo only augments the inputs and modifies the loss computations, which add negligible time to training, we note that the performance improvements from it are obtained nearly for free. Yet, the percentage gain, however, is similar to improvements obtained by each successive state-of-the-art: For example, the improvement of Lite-Mono over HR-Depth , and that of HR-Depth over PackNet  for monocular depth estimation; similarly, the improvement of FusionNet over VOICED and KBNet over FusionNet for depth completion.

\section{Discussion}
\label{sec:discussion}
Conventionally, data augmentation aims to seek visual invariance and create a collection of equivalent classes, i.e., identifying an image and its augmented variant as the same. For geometric tasks, the underlying equivalence is in the 3D scene structures under various illumination conditions, camera, viewpoints, occlusion, etc. Assuming a rigid scene, the shapes populating it should persist regardless of the nuisance variables. This motivates the use of geometric augmentations, as it simulates the nuisances within the image. 
However, adoption of geometric augmentations in unsupervised geometric tasks are obstructed by artifacts introduced during transformations (see Sec. \suppmatref{D} of Supp. Mat.). AugUndo lifts this obstacle by ``undo-ing'' the augmentations before computing the loss.

While AugUndo enables scaling up augmentations for unsupervised training, i.e., depth completion, it may also be applicable for supervised methods; though, we posit that the gain to be less as the artifacts induced from photometric and geometric augmentations of an image tend to be larger than those of a piece-wise smooth depth map. AugUndo is also limited to 2D augmentations; whereas, nuisances modeled by it are projections of the 3D scene (see limitations in Sec. \suppmatref{M} in Supp. Mat.). We leave extensions to 3D for future work. We also only consider a single image and sparse depth map as input. Likewise, extensions can be made towards multi-frame tasks such as stereo, optical flow, pose estimation, etc., but one must account for their specifics and problem setups, i.e., stereo assumes frontoparallel views. This is outside of our scope, so we leave them as future directions. This work paves way for the empirical success in unsupervised geometric tasks that we have observed in other visual recognition tasks. 

\textbf{Acknowledgements.} This work was supported by was supported by NSF 2112562 Athena AI Institute.

\bibliographystyle{ECCV2024/splncs04}
\bibliography{references}

\clearpage
\maketitlesupplementary
\appendix

\textbf{Summary.} In \secref{sec:implementation_details}, we provide implementation details and  hyperparameters, such as learning rate schedule and crop size, used to reproduce our results. In \ref{sec:augmentation_parameters}, we provide augmentation parameters used to train each method on each dataset. \secref{sec:evaluation_metrics} lists the evaluation metrics used in our quantitative results. To illustrate the motivation behind AugUndo, we provide examples of image and sparse depth degradation through typical augmentation routines and add an extended discussion in \secref{sec:extended_motivation}. We provide details on datasets used in our experiments in \secref{sec:datasets}. In \secref{sec:augundo_algo}, we provide a detail walk-through of the AugUndo algorithm. To study the contribution of each of the proposed augmentation, we present a comprehensive ablation study in \secref{sec:ablation_completion}. We further extend the sensitivity study to sparsity from the main paper conducted on VOID to VOID500 and VOID150 in \secref{sec:sensitivity_depth_completion}. We present our results on zero-shot generalization from KITTI to Waymo (depth completion) and Make3d (monocular depth estimation) in \secref{sec:zero_shot_kitti}. We also show a sensitivity study on the effect of different augmentations in \secref{sec:sensitivity_hyperparameters}. In \secref{sec:additional_results_depth_completion}, we include additional results for depth completion, including extensive quantitative results such as evaluations on MonDi and DesNet, comparing modeling AugUndo as change in camera pose and parameters, and the effects of naively applying geometric augmentations during unsupervised training. In \secref{sec:additional_results_monocular_depth}, we show additional results on monocular depth estimation. Finally, we discuss limitations in \secref{sec:limitations}. %

\section{Implementation details}
\label{sec:implementation_details}
For unsupervised depth completion, we implemented our method in PyTorch and incorporated our augmentation pipeline into the codebases of VOICED \cite{wong2020unsupervised}, FusionNet \cite{wong2021learning}, KBNet \cite{wong2021unsupervised}, MonDi \cite{liu2022monitored}, and DesNet \cite{yan2023desnet}. For monocular depth estimation, we implement our method in PyTorch according to \cite{godard2019digging}. Specifically, we implemented our augmentation pipeline into the codebases of Monodepth2 \cite{godard2019digging}, HR-Depth \cite{lyu2021hr}, and Lite-Mono \cite{zhang2023lite}. Details of each task are described below.

\textit{Unsupervised depth completion.} The models are optimized using Adam \cite{kingma2015adam} with $\beta_1 = 0.9$ and $\beta_2 = 0.999$. For VOID: We used an input batch size of 12 with random crop size of $416 \times 512$ for KBNet and DesNet, a batch size of 8 without cropping for FusionNet and VOICED, and a batch size of 8 with random crop size of $448 \times 576$ for MonDi. We trained each KBNet and DesNet for 40 epochs with base learning rate of $1 \times 10^{-4}$ for 20 epochs and decreased to $5 \times 10^{-5}$ for the last 20 epochs. We trained FusionNet and VOICED for 20 epochs with a base learning rate of $1 \times 10^{-4}$ for 10 epochs and decreased to $5 \times 10^{-5}$ for the last 10 epochs. For KITTI: We used a batch size of 8 and random crop size of 320 by 768 for all models. We trained KBNet for 60 epochs, with $5 \times 10^{-5}$ for 2 epochs,  $1 \times 10^{-4}$ until 8th epoch, $2 \times 10^{-4}$ until 30th epoch, $1 \times 10^{-4}$ until the 45th epoch and $5 \times 10^{-5}$ until the 60th epoch. We trained FusionNet for 30 epochs $2 \times 10^{-4}$ for 16 epochs, $1 \times 10^{-4}$ until 24th epoch and $5 \times 10^{-5}$ until the 30th epoch. We trained VOICED for 30 epochs $2 \times 10^{-4}$ for 16 epochs, $6 \times 10^{-5}$ until 24th epoch and $3 \times 10^{-5}$ until the 30th epoch.

We ensure that all baseline methods can reproduce or exceed the numbers originally reported by the authors. Since the authors of DesNet did not publish their code implementation, we re-implemented their method to the best of our ability. All reported results of AugUndo are based on those same settings with the exception of the augmentation scheme.

\textit{Unsupervised monocular depth estimation.} The models are optimized using Adam \cite{kingma2015adam} with $\beta_1 = 0.9$ and $\beta_2 = 0.999$. For VOID, we used a input batch size of 12 and random crop size of $256 \times 448$ for all models. For Monodepth2, we trained with learning rate $1\times 10^{-8}$ for $2$ epochs, $1\times 10^{-5}$ for the next 23 epochs, and $1\times 10^{-6}$ for the next $25$ epochs. For HR-Depth, we train with  learning rate $1\time 10^{-4}$ for 15 epochs, and $1\times 10^{-5}$ for the final 5 epochs. For Lite-Mono, we train with learning rate $5\times 10^{-4}$ for 35 epochs. For KITTI dataset, We used a input batch size of 12 and a random crop size of $192 \times 640$. We trained the models for 20 epochs with an initial learning rate of $1 \times 10^{-4}$ and drop the learning rate to $1 \times 10^{-5}$ at the 15th epoch. The smoothness loss weight for KITTI is set to 0.001 as per \cite{godard2019digging} and 0.01 for VOID as VOID contains more indoor scenes with homogeneous surfaces. 

We ensure that all baseline methods can reproduce or exceed the numbers originally reported by the authors. All reported results of AugUndo are based on those same settings with the exception of the augmentation scheme. Note: For Monodepth2 and HR-Depth, we initialize the ResNet encoder weight with the Imagenet-pretrained weight downloaded from PyTorch website, as specified in their Github repository. However, we cannot locate the pretrained weight used for Lite-Mono throughout their repository, which left us no choice but to train their model from scratch. Nonetheless, this does not affect the validity of the study as the comparison is made between models initialized from scratch where the difference is only in the augmentations schemes: standard convention used in the original papers and repositories, and AugUndo.

\section{Augmentations}
\label{sec:augmentation_parameters}
Through a search over augmentation types and values, we found a consistent set of augmentations that tends to yield improvements across all methods with small changes to degree of augmentation catered to each method. All augmentations are applied with a 50\% probability. We note that performance gain can be obtained by typical set and ranges of augmentations and does not require a meticulous selection of hyper-parameters (see \figref{fig:degree_aug}).

\textit{VOID.} 
For depth completion, we applied photometric transformations of random brightness from 0.5 to 1.5, contrast from 0.5 to 1.5, saturation from 0.5 to 1.5, and hue from -0.1 to 0.1. We applied image patch removal by selecting between 0.1\% to 0.5\% of the pixels and removing 5 $\times$ 5 patches centered on them -- approximately removing between 2.5\% to 12.5\% of the image. We applied random sparse depth point removal at a rate between 60\% and 70\% of all sparse points. We applied geometric transformations of random horizontal and vertical flips, up to 10\% of translation and between -25 to 25 degrees of rotation. For KBNet and FusionNet, we applied random resize factor between 0.6 to 1.1, while for VOICED, we used 0.7 to 1.1 because the scaffolding step of VOICED requires at least 3 points, but as discussed above, resampling causes loss and factors smaller than 0.7 often cause all points to be dropped. 
For monocular depth estimation, we applied photometric transformations of random brightness from 0.5 to 1.5, contrast from 0.5 to 1.5, saturation from 0.5 to 1.5, hue from -0.1 to 0.1. We applied geometric transformation of random rotation between -10 to 10 degrees and random horizontal flipping. We further applied random resize factor between 0.8 to 1. 

\textit{KITTI.} For depth completion, we applied random brightness, contrast, saturation from 0.5 to 1.5 and random hue from -0.1 to 0.1. We applied image patch removal by selecting between 0.1\% to 0.5\% of the pixels and removing 5 $\times$ 5 patches centered on them. We applied random sparse depth point removal at a rate between 60\% and 70\% of all sparse points. We further applied random horizontal flips, up to 10\% of translation, resizing factors between 0.8 to 1.2, and between -20 to 20 degrees of rotation. We found that vertical flips are detrimental to performance. For monocular depth estimation, we applied random brightness, contrast, saturation from 0.5 to 1.5 and random hue from -0.1 to 0.1. We applied a random rotation between -30 to 30 degrees and a random translation of up to 30\% of the image. We further apply horizontal flip to the image.

\begin{table}[h]
\centering
    \caption{
        \textit{Error metrics for depth completion and monocular depth estimation.} $d_{gt}$ denotes ground truth depth and evaluated where values are available for a given image. 
    }
    \scalebox{0.8}{
    \begin{tabular}{l l}
        \midrule
            Metric & Definition \\ \midrule
            MAE &$\frac{1}{|\Omega|} \sum_{x\in\Omega} |\hat d(x) - d_{gt}(x)|$ \\
            RMSE & $\big(\frac{1}{|\Omega|}\sum_{x\in\Omega}|\hat d(x) - d_{gt}(x)|^2 \big)^{1/2}$ \\
            iMAE & $\frac{1}{|\Omega|} \sum_{x\in\Omega} |1/ \hat d(x) - 1/d_{gt}(x)|$ \\
            iRMSE& $\big(\frac{1}{|\Omega|}\sum_{x\in\Omega}|1 / \hat d(x) - 1/d_{gt}(x)|^2\big)^{1/2}$ \\ 
            AbsRel &$\frac{1}{|\Omega|} \sum_{x\in\Omega} \frac{|\hat d(x) - d_{gt}(x)|}{d_{gt}(x)}$ \\
            SqRel &$\frac{1}{|\Omega|} \sum_{x\in\Omega} \frac{|\hat d(x) - d_{gt}(x)|^2}{d_{gt}(x)}$ \\
            Accuracy & \% of $ z(x)$ s.t. $\delta \doteq \max (\frac{z(x)}{z_{gt}(x)}, \frac{z_{gt}(x)}{z(x)}) < $ threshold \\
            \midrule
        \end{tabular}
    }
\label{tab:error_metrics_depth_completion}
\end{table}

\section{Evaluation metrics}
\label{sec:evaluation_metrics}
The evaluation metrics used for depth completion and monocular depth estimation are shown in \tabref{tab:error_metrics_depth_completion}. Depth completion models are evaluated with MAE, RMSE, iMAE, iRMSE. Monocular depth estimation models are evaluated with MAE, RMSE, AbsRel, SqRel, and accuracy ($\delta < 1.25$, $\delta < 1.25^2$, $\delta < 1.25^3$).  We note that as monocular depth is inferred, at most, up to an unknown scale, we perform scale matching during evaluation by using median scaling with respect to the ground truth before computing each error or accuracy metric.

\begin{figure*}[t]
    \centering
    \includegraphics[width=\textwidth]{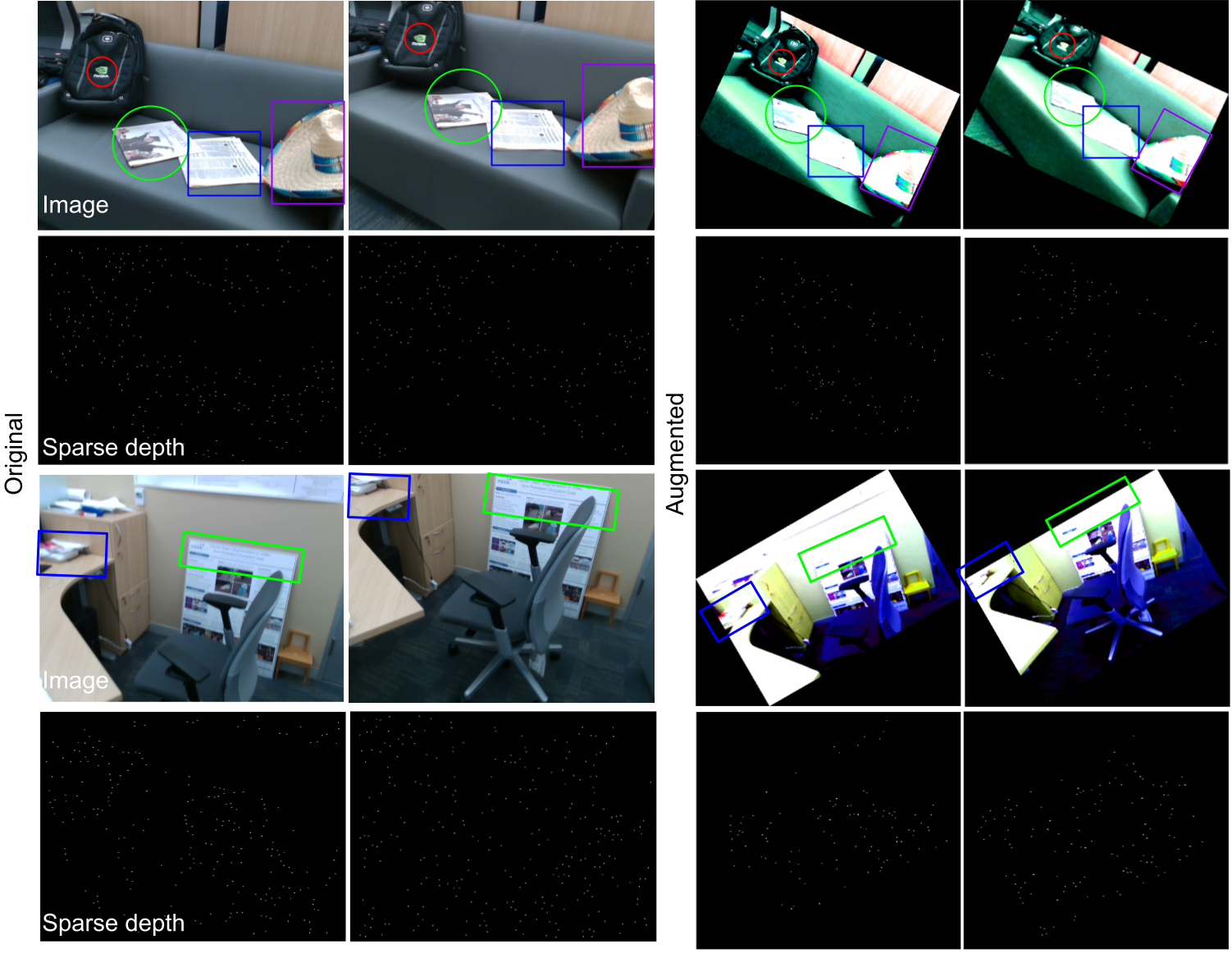}
    \caption{\textit{Motivation.} After augmenting the inputs with photometric and geometric transformations, the previously exciting image textures (for establishing correspondence across images) and sparse points in the original inputs, that would have served as supervision, are now largely saturated in intensity and homogeneous, and lost due to resampling, respectively (best viewed in 2$\times$).}
    \label{fig:motivation}
\end{figure*}

\section{Extended Discussion of Motivation}
\label{sec:extended_motivation}
In the main paper, we discussed the motivation behind the approach. Here, we provide an extended discussion: Training unsupervised depth completion methods \cite{yang2019dense,wong2020unsupervised,wong2021adaptive,wong2021learning,wong2021unsupervised} relies on a photometric reconstruction term, sparse depth reconstruction term, and a (generic) regularizer, such as local smoothness; a related problem, unsupervised monocular depth estimation \cite{godard2019digging,fei2019geo,lao2024depth,wong2019bilateral,wong2020targeted,zeng2024wordepth}, omits the sparse depth term. The photometric reconstruction term constraints the solution in regions where there are sufficiently exciting textures and co-visible between images such as those highlighted in \figref{fig:motivation}.
The sparse depth reconstruction term constraints the solution anywhere with a sparse point. Everywhere else in the image are inherently ambiguous, so we must rely on the regularizer. \figref{fig:motivation} shows that after one has applied a number of photometric and geometric transformations, the previously exciting textures that would have served as our supervision are now largely saturated in intensity and homogenous. The number of points in the sparse depth maps have also been greatly reduced.

In conventional augmentation scheme for \textit{other} (semantic) vision tasks, i.e. classification, detection, and segmentation, it is typical for one to introduce saturation in image intensities, block artifacts from interpolation and loss during resampling. These are some of the side-effects that are \textit{desirable} in the aforementioned tasks: introducing these nuisance variabilities enables the model to learn to be invariant to them (i.e., to learn them away), which yields more generalizable and robust representations. This is also true for supervised depth completion \cite{ezhov2024all,van2019sparse,park2020non,singh2023depth} and estimation methods \cite{eigen2015predicting,yin2019enforcing,ranftl2020towards,ranftl2021vision,upadhyay2023enhancing}. However, this is not the case for learning unsupervised depth. Because the supervision signal comes from reconstructing the input image and sparse depth map, the more we augment the data (causing a loss of photometric correspondences across image frames and a loss of points in the sparse depth map), the more we degrade the supervision signal (see \cref{tab:naive_augmentations_depth_completion,tab:naive_augmentations_depth_estimation} in \secref{sec:additional_results_depth_completion} for empirical evidence of this phenomenon). So, it is not too surprising that conventional augmentation procedures, both photometric and geometric, have seen limited use beyond small changes in image intensities, and flipping.

Nevertheless, photometric augmentations help model the diverse range of illumination conditions and colors of object that may populate the scene. Geometric augmentations can simulate the various camera parameters and motion, for example, image translation can simulate principle point offsets or it can approximate small baseline movements, image resizing with cropping can simulate different focal lengths (zooming) or it can model camera movements, while in-plane rotations can model camera orientation. These augmentations are often viewed as essential to training pipelines for other vision tasks, but are detrimental for unsupervised depth completion; yet, without them, one may encounter robustness and generalization issues. As the root of the problem lies in the supervision signal, we investigate an approach to ``undo'' the augmentations during (or right before) the loss computation step. Doing so enables one to feed forward augmented inputs, but compute the loss on the original inputs with no loss in the training signal. Extensive experiments show that our approach improves performance and zero-shot generalization across a number of methods for both indoor and outdoor scenarios.

\section{Datasets}
\label{sec:datasets}
We conduct experiments on six datasets (two for training and four for generalization) in total. Details of each dataset are provided below.

\textbf{KITTI} \cite{geiger2013vision} contains 61 driving scenes with research in autonomous driving and computer vision. It contains calibrated RGB images with sychronized point clouds from Velodyne lidar, inertial, GPS information, etc. For depth completion \cite{uhrig2017sparsity}, there are $\approx$80,000 raw image frames and associated sparse depth maps, each with a density of $\approx$5\%. Ground-truth depth is obtained by accumulating 11 neighbouring raw lidar scans. Semi-dense depth is available for the lower $30\%$ of the image space. We test on the official validation set of 1,000 samples because the online test server has submission restrictions to accomodate multiple trials. For depth estimation, we used Eigen split \cite{eigen2015predicting}, following \cite{zhou2017unsupervised} to preprocess and remove static frames. The remaining training set contains 39,810 monocular triplets and the validation set contains 4,424 triplets. The testing set contains 697 monocular images. We follow the evaluation protocol of \cite{uhrig2017sparsity, eigen2014depth}, where output depth is evaluated where ground truth exists between 0 to 80.0 meters. %

\textbf{VOID} \cite{wong2020unsupervised} comprises indoor (laboratories, classrooms) and outdoor (gardens) scenes with synchronized $640 \times 480$ RGB images and sparse depth maps. XIVO \cite{fei2019geo}, a VIO system, is used to obtain the sparse depth maps that contain approximately 1500 sparse depth points with a density of about 0.5\%. Active stereo is used to acquire the dense ground-truth depth maps. In contrast to the typically planar motion in KITTI, VOID has 56 sequences with challenging 6 DoF motion captured on rolling shutter. 48 sequences (about 45,000 frames) are assigned for training and 8 for testing (800 frames). We follow the evaluation protocol of \cite{wong2020unsupervised} where output depth is evaluated where ground truth exists between 0.2 and 5.0 meters.

\textbf{NYUv2} \cite{silberman2012indoor} consists of 372K synchronized $640 \times 480$ RGB images and depth maps for 464 indoors scenes (household, offices, commercial), captured with a Microsoft Kinect. The official split consisting in 249 training and 215 test scenes. We use the official test set of 654 images. Because there are no sparse depth maps provided, we sampled  $\approx 1500$ points from the depth map via Harris corner detector \cite{harris1988combined} to mimic the sparse depth produced by SLAM/VIO. We test models trained on VOID to evaluate their generalization to NYUv2. We follow the evaluation protocol of \cite{wong2020unsupervised} where output depth is evaluated where ground truth exists between 0.2 and 5.0 meters.

\textbf{ScanNet} \cite{dai2017scannet} consists of RGB-D data for 1,513 indoor scenes with 2.5 million images and corresponding dense depth map. Because there are no sparse depth maps provided, we sampled  $\approx 1500$ points from the depth map via Harris corner detector \cite{harris1988combined} to mimic the sparse depth produced by SLAM/VIO. We followed \cite{dai2017scannet} and used 100 scenes (scene707-scene806), for zero-shot generalization for models trained on VOID. The output depth is evaluated where ground truth exists between 0.2 and 5.0 meters.

\textbf{Waymo Open Dataset}~\cite{sun2020scalability} contains 1920$\times$1280 RGB images and lidar scans from autonomous vehicles. The training set contains $\approx$158K images from 798 scenes and the validation set $\approx$40K images from 202 scenes, collected at 10Hz. Objects are annotated across the full 360$^{\circ}$ field. Sparse depth maps are obtained by reprojecting the point cloud scan from the top lidar to the camera frame. Ground truth is obtained by reprojecting both front facing lidars as well as those collected 10 time steps forward and backwards (approximately 1 second of capture) to a given camera frame at a specific time step to densify the sparse depth. We used the object annotations to remove all moving objects to ensure that reprojected points respects the static scene assumption. We also performed outlier removal to filter out errorenous (noisy) points. The output depth is evaluated where ground truth exists between a 1.5 and 80.0 meters range.

\textbf{Make3d} \cite{saxena2008make3d} contains 134 test images with $2272 \times 1707$ resolution. Ground-truth depth maps are given at $305 \times 55$ resolution and must be rescaled and interpolated. We use the central cropping proposed by \cite{godard2017unsupervised} to get a $852 \times 1707$ center crop of the image. We use standard Make3d evaluation protocol and metrics. We use Make3d to test the generalization of monocular depth estimation models trained on KITTI.

\begin{algorithm*}[t]
	\begin{algorithmic}[1]
		\REQUIRE{Depth completion network $f_\theta$, Images $I_t, I_\tau$, Sparse depth $z_t$, \\ Relative pose $g_{rt}$, Intrinsics $K$}
            \STATE Sample $\{ T_{pt, I}^1 \ldots T_{pt, I}^k \}$ from $ T_{pt, I}^i \in \mathcal{A}_{pt, I}$, and compose  \\ 
            \ \ \ \ $T_{pt, I} = T_{pt, I}^1 \circ T_{pt, I}^2 \circ \cdots \circ T_{pt, I}^k$
            \STATE Sample $\{ T_{pt, z}^1 \ldots T_{pt, z}^j \}$ from $ T_{pt, z}^i \in \mathcal{A}_{pt, z}$, and compose  \\ 
            \ \ \ \ $T_{pt, z} = T_{pt, z}^1 \circ T_{pt, z}^2 \circ \cdots \circ T_{pt, z}^j$
            \STATE Sample $\{ T_{ge}^1 \ldots T_{ge}^m \}$ from $ T_{ge}^i \in \mathcal{A}_{ge}$, and compose  \\ 
            \ \ \ \ $T_{ge} = T_{ge}^1 \circ T_{ge}^2 \circ \cdots \circ T_{ge}^m$
            \STATE Compose the inverse geometric transform  \\ 
            \ \ \ \ $T_{ge}^{-1} = ({T_{ge}^m})^{-1} \circ (T_{ge}^{m-1})^{-1} \circ \cdots \circ ({T_{ge}^1})^{-1}$
            \STATE Compute the coordinates after geometric transform  \\ 
            \ \ \ \ $\begin{bmatrix}x' & 1 \end{bmatrix}^\top = T_{ge} \begin{bmatrix}x & 1\end{bmatrix}^{\top}$ (Eqn. \mainpaperref{3} from main paper)
            \STATE Augment $I_t$ with photometric and geometric transformations \\ 
            \ \ \ \ $I_t'(x') = T_{pt,I}(I_t)(x)$ (Eqn. \mainpaperref{4} from main paper)
            \STATE Augment $z_t$ with occlusion and geometric transformations  \\ 
            \ \ \ \ $z_t'(x') = T_{pt,z}(z_t)(x)$ (Eqn. \mainpaperref{4} from main paper)
            \STATE Obtain depth prediction $\hat{d}'_t = f_\theta(I'_t, z'_t)$
            \STATE Compute coordinates of the inverse geometric transformation  \\ 
            \ \ \ \ $\begin{bmatrix} x'' & 1\end{bmatrix}^\top = T^{-1}_{ge} \begin{bmatrix}x' & 1 \end{bmatrix}^{\top}$ (Eqn. \mainpaperref{5} from main paper)
            \STATE Apply inverse geometric transformation on output depth map:  \\ 
            \ \ \ \ $\hat{d}_t(x'') = \hat{d}_t'(x')$ (Eqn. \mainpaperref{6} from main paper)
            \STATE Reconstruct $I_t$ from $I_\tau$ using Eqn. \mainpaperref{1} from main paper, i.e., $\hat{I}_{t\tau} = I_\tau (\pi g_{\tau t} K^{-1} \bar{x} \hat{d}_{t})$ 
            \STATE Minimize reconstruction losses between $\hat{I}_{t\tau}$ and $I_t$, and 
            $\hat{d}_t$ and $z_t$, and the regularizer (Eqn. \mainpaperref{2} from main paper)
	\end{algorithmic}
	\caption{\textsc{AugUndo}}
	\label{alg:main-algorithm} 
 
\end{algorithm*}

\section{The AugUndo Algorithm}
\label{sec:augundo_algo}
We assume that we are given (i) $I : \Omega \subset \mathbb{R}^2 \rightarrow \mathbb{R}^3_+$ an RGB image $I_t$, (ii), its associated sparse point cloud projected onto as a depth map $z : \Omega_z \subset \Omega \rightarrow \mathbb{R}_+$, $z_t$, (iii) camera intrinsic calibration matrix $K \in \mathbb{R}^{3 \times 3}$, (iv) a sequence of images $I_\tau$ for $\tau \in \{t-1, t+1\}$ during training, and (v) the relative pose $g_{\tau t}$ between the image frame $I_t$ and some temporally adjacent image $I_\tau$. Given a image and its associated sparse depth map, a depth completion network learns a map the inputs to the output depth map $\hat{d}_t := f_\theta(I_t, z_t) \in \mathbb{R}_+^{H \times W}$. In the main paper, we denoted photometric and occlusion augmentations as $\mathcal{A}_{pt}$ for ease of notation. Here, for specificity, we define $\mathcal{A}_{pt,I}$ as the set of possible photometric (including occlusion) transformations for the image,  $\mathcal{A}_{pt, z}$ as the possible set of occlusion augmentations for sparse depth maps, and $\mathcal{A}_{ge}$ as the possible set of all geometric transformations. We additionally assumes we have the set of geometric transformations $T_{ge}$ used during augmentation and their inverse transformations $T^{-1}_{ge}$. \algref{alg:main-algorithm} is the procedural algorithm of AugUndo and details the step by step augmentation and loss computation pipelines, given our inputs.

\begin{table}[t]
  \centering
      \caption{\textit{Ablation study for KBNet on VOID dataset}. BRI stands for brightness, CON for contrast, SAT for saturation, FLP for horizontal and vertical flip, TRN for translation, ROT for rotation, RZD for resize down, RZU for resize up, RMP for point removal, RMI for image patch removal. \textbf{Bold} denotes AugUndo, \textit{italicized} the standard augmentation protocol. RMP, FLP, RZD, RZU have the highest influence; only when BRI, CON, SAT, HUE (color jitter) are disabled do they have non-negligible effect. Best results are achieved by using AugUndo.}
    \scalebox{0.77}{
    \begin{tabular}{c c c c c c c c c c c c c c c}
        \midrule
        \multicolumn{11}{c}{Augmentation settings} & \multicolumn{4}{c}{Evaluation metrics} \\
        \cmidrule(lr){1-11} \cmidrule(lr){12-15}
        BRI & CON & SAT & HUE & FLP & TRN & ROT & RZD & RZU & RMP & RMI & MAE & RMSE & iMAE & iRMSE \\ 
        \midrule
        \checkmark & \checkmark & \checkmark & \checkmark & \checkmark &  \checkmark & \checkmark & \checkmark & \checkmark & \checkmark & \checkmark
        & \textbf{33.32}$\pm$0.18 & \textbf{85.67}$\pm$0.39 & \textbf{16.61}$\pm$0.29 & \textbf{41.24}$\pm$0.60 \\
        \checkmark & \checkmark & \checkmark & \checkmark & \checkmark & & & & & &
        & \textit{38.11}$\pm$0.77 & \textit{95.22}$\pm$1.72 & \textit{19.51}$\pm$0.14 & \textit{46.70}$\pm$0.48 \\
        & \checkmark & \checkmark & \checkmark & \checkmark & \checkmark & \checkmark & \checkmark & \checkmark & \checkmark & \checkmark
        & 33.46$\pm$0.27 & 86.02$\pm$0.69 & 16.84$\pm$0.22 & 41.78$\pm$0.63 \\
        \checkmark & & \checkmark & \checkmark & \checkmark & \checkmark & \checkmark & \checkmark & \checkmark & \checkmark & \checkmark
        & 33.79$\pm$0.09 & 86.38$\pm$0.36 & 17.12$\pm$0.19 & 42.12$\pm$0.54 \\
        \checkmark & \checkmark & & \checkmark & \checkmark & \checkmark & \checkmark & \checkmark & \checkmark & \checkmark & \checkmark
        & 33.73$\pm$0.30 & 85.84$\pm$0.22 & 17.05$\pm$0.16 & 41.73$\pm$0.47 \\
        \checkmark & \checkmark & \checkmark &  & \checkmark & \checkmark & \checkmark & \checkmark & \checkmark & \checkmark & \checkmark
        & 33.60$\pm$0.16 & 86.47$\pm$0.69 & 16.95$\pm$0.27 & 41.65$\pm$0.44 \\
        & & & & \checkmark & \checkmark & \checkmark & \checkmark & \checkmark & \checkmark & \checkmark
        & 34.14$\pm$0.37 & 87.26$\pm$0.74 & 17.09$\pm$0.17 & 42.57$\pm$1.33 \\
        \checkmark & \checkmark & \checkmark & \checkmark & & \checkmark & \checkmark & \checkmark & \checkmark & \checkmark & \checkmark
        & 44.38$\pm$0.89 & 106.87$\pm$0.89 & 22.91$\pm$0.68 & 52.55$\pm$0.57 \\
        \checkmark & \checkmark & \checkmark & \checkmark & \checkmark & & \checkmark & \checkmark & \checkmark & \checkmark & \checkmark
        & 33.92$\pm$0.19 & 87.19$\pm$0.55 & 17.09$\pm$0.03 & 42.23$\pm$0.23 \\
        \checkmark & \checkmark & \checkmark & \checkmark & \checkmark & \checkmark & & \checkmark & \checkmark & \checkmark & \checkmark
        & 33.68$\pm$0.19 & 85.86$\pm$0.51 & 16.92$\pm$0.03 & 41.47$\pm$0.18 \\
        \checkmark & \checkmark & \checkmark & \checkmark & \checkmark & \checkmark & \checkmark &  & \checkmark & \checkmark & \checkmark
        & 35.77$\pm$0.43 & 89.88$\pm$0.84 & 17.81$\pm$0.23 & 42.75$\pm$0.38 \\
        \checkmark & \checkmark & \checkmark & \checkmark & \checkmark & \checkmark & \checkmark & \checkmark & & \checkmark & \checkmark				
        & 35.69$\pm$0.25 & 90.40$\pm$0.71 & 17.75$\pm$0.12 & 42.82$\pm$0.20 \\
        \checkmark & \checkmark & \checkmark & \checkmark & \checkmark & \checkmark & \checkmark & \checkmark & \checkmark & & \checkmark
        & 37.73$\pm$0.27 & 92.62$\pm$0.20 & 19.44$\pm$0.18 & 45.47$\pm$0.37 \\
        \checkmark & \checkmark & \checkmark & \checkmark & \checkmark & \checkmark & \checkmark & \checkmark & \checkmark & \checkmark &
        & 33.64$\pm$0.22 & 86.26$\pm$0.23 & 16.80$\pm$0.13 & 41.60$\pm$0.64 \\
        \midrule
    \end{tabular}}
\label{tab:kbnet_ablation}
\end{table}

\section{Ablation Study for Depth Completion}
\label{sec:ablation_completion}

In the main paper, we demonstrated AugUndo for unsupervised depth completion methods on VOID1500 and VOID500 benchmarks using the most performant set of augmentations we found. Here, we provide an ablation study for each of the augmentations used to study their individual contributions.

\tabref{tab:kbnet_ablation} shows a comprehensive ablation study following augmentation settings above. From our best reported settings on KBNet, we removed individual augmentations to show their empirical contribution. Note: random brightness, contrast, saturation, and hue (BRI, SAT, CON, HUE) together is the standard color jitter employed by all current methods. We also test removing all color jitter augmentations to further quantify its effect. 

We found that random flips (FLP) has the largest impact of all of the augmentations -- increasing the error by an average of 30\%; this is because it simulates different scene layouts. As there are no viable intensity augmentations for sparse depth, RMP is the only one to explicitly increase variability in the data modality hence it also has large influence. We note that other geometric augmentations contribute as well, i.e., rotation, resizing, and translation, to discard points. By computing the loss on the original inputs, we reconstruct the removed points, which in fact serves as an additional training signal to map RGB to depth.

Photometric augmentations, individually, have small effect on performance. To get a non-neglible effect, one must disable all color jitter (\tabref{tab:kbnet_ablation} , row 7), which still yielded small increases in errors. This shows the limitations of existing augmentations, which relies heavily on color jittering. We note that we use a larger value range in color jitter than existing works, so the impact is expected to be even smaller for existing works. Admittedly, the most important one (FLP) is currently being used by all methods, but resize, image patch and point removal, play a large role. The best results are obtained when using all of the proposed augmentations, demonstrating the importance of scaling up both photometric and geometric augmentations.

\begin{table}[t]
    \centering
        \caption{
        \textit{Sensitivity study of depth completion on VOID.} Reported scores are mean and standard deviation over four independent trials. 
        We compare the sensitivity of models trained on VOID1500, with standard augmentations and AugUndo, by testing them on VOID500 and VOID150. Sparse depth maps within VOID1500 contains approximately 1500 points, and those within VOID500 and VOID150 contain approximately 3$\times$ and 10$\times$ less, respectively. AugUndo improves performance by an average of %
        19.66\% and 19.62\% across all unsupervised methods and evaluation metrics on VOID500 and VOID150, respectively. For distillation method (marked by *), MonDi, AugUndo improves by an average of 9.02\% and 10.25\% on VOID500 and VOID150, respectively. Despite MonDi is supervised by pseudo ground truth in addition to typical unsupervised losses, AugUndo can still boost performance. 
    }
    \setlength\tabcolsep{1pt}
    \scalebox{0.7}{\begin{tabular}{l c c c c c c c c}
        \midrule
         & \multicolumn{4}{c}{VOID500} & \multicolumn{4}{c}{VOID150} \\
         \cmidrule(lr){2-5} \cmidrule(lr){6-9}
         Method & MAE $\downarrow$& RMSE $\downarrow$& iMAE $\downarrow$& iRMSE $\downarrow$ & MAE $\downarrow$& RMSE $\downarrow$& iMAE $\downarrow$& iRMSE $\downarrow$ \\ 
        \midrule
        VOICED
        & 137.01$\pm$4.23 & 235.80$\pm$7.82 & 71.36$\pm$1.86 & 130.63$\pm$5.66 
        & 209.59$\pm$5.18 & 329.71$\pm$10.01 & 130.45$\pm$5.63 & 229.79$\pm$14.09
        \\
        + AugUndo
        & \textbf{92.99}$\pm$1.11 & \textbf{176.94}$\pm$1.38 & \textbf{46.43}$\pm$0.85 & \textbf{91.10}$\pm$1.64
        & \textbf{151.77}$\pm$1.99 & \textbf{262.61}$\pm$2.19 & \textbf{88.65}$\pm$0.76 & \textbf{169.29}$\pm$2.71 
        \\ 
        \midrule
        FusionNet
        & 97.73$\pm$0.73 & 194.32$\pm$1.36 & 58.65$\pm$1.31 & 122.95$\pm$3.04
        & 158.03$\pm$1.97 & 284.23$\pm$3.05 & 113.67$\pm$2.03 & 223.41$\pm$0.93 
        \\
        + AugUndo
        & \textbf{74.97}$\pm$1.15 & \textbf{162.71}$\pm$1.86 & \textbf{40.44}$\pm$1.09 & \textbf{92.11}$\pm$1.01 
        & \textbf{126.16}$\pm$1.44 & \textbf{246.16}$\pm$2.46 & \textbf{86.13}$\pm$4.46 & \textbf{181.08}$\pm$8.00 
        \\
        \midrule
        KBNet        
        & 78.44$\pm$1.39 & 178.17$\pm$3.27 & 37.56$\pm$0.61 & 83.43$\pm$1.89 
        & 149.13$\pm$3.29 & 306.30$\pm$8.74 & 70.74$\pm$2.26 & 136.75$\pm$5.55 		
        \\			
        + AugUndo
        & \textbf{66.97}$\pm$0.81 & \textbf{151.55}$\pm$2.03 & \textbf{31.63}$\pm$0.53 & \textbf{71.90}$\pm$0.82
        & \textbf{117.16}$\pm$4.51 & \textbf{239.60}$\pm$10.96 & \textbf{57.65}$\pm$1.47 & \textbf{112.81}$\pm$1.75 
        \\
        \midrule
        DesNet
        & 74.89$\pm$0.67 & 170.32$\pm$2.03 & 35.62$\pm$0.72 & 78.30$\pm$1.31 
        & 139.54$\pm$4.46 & 287.95$\pm$12.69 & 65.00$\pm$1.24 & 123.81$\pm$2.17
         \\			
        + AugUndo
        & \textbf{67.78}$\pm$1.10 & \textbf{153.46}$\pm$2.64 & \textbf{32.09}$\pm$0.48 & \textbf{71.96}$\pm$1.05
        & \textbf{117.93}$\pm$3.77 & \textbf{239.49}$\pm$8.44 & \textbf{58.13}$\pm$1.14 & \textbf{112.78}$\pm$2.33
        \\
        \midrule
        \midrule
        MonDi*   		
        & 73.90$\pm$0.61 & 191.69$\pm$2.64 & 31.78$\pm$1.52 & 72.67$\pm$2.17
        & 146.33$\pm$5.57 & 343.25$\pm$13.70 & 60.43$\pm$3.54 & 114.18$\pm$2.21 		
        \\			
        + AugUndo
        & \textbf{69.90}$\pm$3.05 & \textbf{157.02}$\pm$7.46 & \textbf{29.61}$\pm$0.12 & \textbf{68.48}$\pm$0.57
        & \textbf{127.08}$\pm$4.51 & \textbf{299.42}$\pm$10.96 & \textbf{54.47}$\pm$1.47 & \textbf{108.23}$\pm$1.75 
        \\
        \midrule
    \end{tabular}}
\label{tab:depth_completion_void_sensitivity_sparsity}
\end{table}

\section{Sensitivity Study for Depth Completion}
\label{sec:sensitivity_depth_completion}
Given that conventional augmentation pipelines are not applied towards sparse depth modality, it is possible that a model will overfit to the sparse point cloud, which describes the coarse 3D scene structure. Overfitting to scenes, hence, can limit generalization and increase sensitivity to the configuration of sparse points. In the main paper, we presented results for depth completion on VOID1500 and VOID500. Here, we test the effect of AugUndo on various sparse depth input densities. We presented results on VOID500 (repeated for side-by-side comparison) and VOID150 in \tabref{tab:depth_completion_void_sensitivity_sparsity} (left and right, respectively). VOID500 contains approximately 500 sparse points per point cloud and VOID150 contains approximately 150 points. All models tested are trained on VOID1500, which contains 1500 sparse points. 

For VOID500 (\tabref{tab:depth_completion_void_sensitivity_sparsity}, left), which is a 3$\times$ reduction in density, AugUndo improves the sensitivity to changes in the sparse points by an average of 19.66\% across all methods, and 30.57\%, 23.92\%,  14.79\%, 9.35\% for VOICED, FusionNet, KBNet, and DesNet, respectively. For an even more challenging scenario, also the closest setup to the density of sparse points tracked by a Simultaneous Localization and Mapping (SLAM) and Visual Inertial Odometry (VIO) system, we consider VOID150 (\tabref{tab:depth_completion_void_sensitivity_sparsity}, right). Here, AugUndo improves 
by an average of 19.62\% across all methods, and 26.57\%, 19.18\%, 19.80\%, and 12.95\% for VOICED, FusionNet, KBNet, and DesNet, respectively. We attribute this to geometric and occlusion augmentations:  translation, resize, rotation and sparse points removal. All of these augmentations not only affect photometry, but also the sparse point cloud where points are dropped due to resampling or explicitly removed, and additionally point cloud orientation is also altered.

We note that AugUndo is also helpful for distillation methods  like MonDi \cite{liu2022monitored}. As mentioned in the main text, we conjecture that artifacts caused from transformation of a piece-wise smooth depth map (in the case of supervised or distillation methods) are less severe than those of image and sparse point clouds. Hence, we were expecting the gains for supervised or distillation methods to be small compared to unsupervised methods. However, as shown in the last row of \tabref{tab:depth_completion_void_sensitivity_sparsity}, we observe a surprisingly nontrivial gain when applying AugUndo to MonDi (marked by *). For VOID500, we improve MonDi by 9.02\% on average across on metrics; for VOID150, we improve by 10.25\%. 

\section{Zero-shot Generalization from KITTI}
\label{sec:zero_shot_kitti}
In the main paper, we demonstrated AugUndo for three unsupervised depth completion (VOICED \cite{wong2020unsupervised}, FusionNet\cite{wong2021learning}, and KBNet \cite{wong2021unsupervised}) and three unsupervised monocular depth estimation (Monodepth2 \cite{godard2019digging}, HR-Depth \cite{lyu2021hr}), and Lite-Mono \cite{zhang2023lite}) methods on the KITTI dataset. Due to space constraints, here, we provide additional results for zero-shot generalization from KITTI to Waymo Open Dataset \cite{sun2020scalability} for depth completion and to Make3d \cite{saxena2008make3d} for monocular depth estimation. Similar to our generalization experiments on indoors (VOID to NYUv2 and ScanNet), we will train on KITTI using the conventional augmentation schemes employed by each respective method and compare the resulting models with those trained using AugUndo.

\begin{table*}[t]
    \centering
        \caption{
        \textit{Zero-shot transfer from KITTI to Waymo for depth completion.} %
        AugUndo improves generalization of models trained on KITTI to Waymo by an average of 13.3\% for all evaluation metrics. We note that the sparse depth maps provided in Waymo is considerably denser than KITTI as they are merged from two front separate lidars; hence, FusionNet, which employs a learned (frozen) densification network performs similarly, i.e., the bias introduced by the same frozen densification network (ScaffNet) is strong enough that FusionNet yields similar results for both conventional augmentation scheme and AugUndo. Nonetheless, we still observe considerable improvements.
    }
    \scriptsize
    \begin{tabular}{l l c c c c}
        \midrule
        Dataset & Method & MAE $\downarrow$ & RMSE $\downarrow$ & iMAE $\downarrow$ & iRMSE $\downarrow$ \\ 
        \midrule
        \multirow{6}*{Waymo} & VOICED 
        & 6781.27$\pm$317.21 & 7734.57$\pm$339.77 & 24.58$\pm$1.45 & 27.87$\pm$1.63 \\
        & + AugUndo 
        & \textbf{5965.07}$\pm$367.47 & \textbf{7029.66}$\pm$509.43 & \textbf{18.54}$\pm$1.83 & \textbf{22.10}$\pm$3.96 \\	
        & FusionNet
        & 530.55$\pm$39.23 & 1734.23$\pm$114.97 & 1.27$\pm$0.12 & 2.82$\pm$0.29 \\			
        & + AugUndo
        & \textbf{512.29}$\pm$8.43 & \textbf{1707.34}$\pm$48.25 & \textbf{1.21}$\pm$0.09 & \textbf{2.75}$\pm$0.18 \\
        & KBNet 
        & 625.00$\pm$12.47 & 2167.74$\pm$92.17 & 1.76$\pm$ 0.13 & 5.46$\pm$0.93 \\
        & + AugUndo 
        & \textbf{541.29}$\pm$15.16 & \textbf{2014.14}$\pm$76.52 & \textbf{1.34}$\pm$0.11 & \textbf{3.43}$\pm$0.30 \\
        \midrule
    \end{tabular}
    \vspace{-0em}
\label{tab:depth_completion_zero_shot_outdoors}
\end{table*}

\begin{table*}[t]
    \centering
        \caption{\textit{Zero-shot transfer from KITTI to Make3d for monocular depth estimation.} All models are trained on KITTI. Note: for Monodepth2, we use the numbers reported by \cite{godard2019digging} and the best trial on KITTI.}
    \setlength\tabcolsep{7pt}
    \fontsize{8pt}{8pt}\selectfont 
    \scalebox{0.75}{\begin{tabular}{l l c c c c c c c c c c c}
        \midrule 
        Dataset & Method & MAE $\downarrow$ & RMSE $\downarrow$& Abs Rel $\downarrow$& Sq Rel $\downarrow$ & $\delta < 1.25 \uparrow$ & $\delta < 1.25^2\uparrow$ & $\delta < 1.25^3\uparrow$ \\ 
        \midrule
        \multirow{6}*{Make3d} & Monodepth2 & - & 7.417 & 0.322 & 3.589 & - & - & - \\
         & + AugUndo & 4.109 & \textbf{6.803} & \textbf{0.272} & \textbf{2.769} & 0.606 & 0.848 & 0.936 \\
        & HR-Depth & 4.136 & 6.505 & 0.281 & 2.484 & 0.562 &  0.839 & 0.938 \\
        & + AugUndo & \textbf{4.023} & \textbf{6.428} & \textbf{0.272} & \textbf{2.393} & \textbf{0.584} & \textbf{0.848} & \textbf{0.94} \\
         & Lite-Mono & 5.116 & 8.061 & 0.358 & 4.676 & 0.511 & 0.793 & 0.91 \\
         & + AugUndo & \textbf{4.728} & \textbf{7.518} & \textbf{0.321} & \textbf{3.887} & \textbf{0.529} & \textbf{0.817} & \textbf{0.926} \\ 
        \midrule
    \end{tabular}}
\label{tab:monocular_depth_prediction_generalization_outdoors}
\end{table*}

We begin by presenting results on depth completion in \tabref{tab:depth_completion_zero_shot_outdoors}. Here, we evaluate VOICED, FusionNet, and KBNet models (trained on KITTI using standard augmentation pipelines and AugUndo) on the Waymo Open Dataset. Overall, training with AugUndo improves existing methods by an average 13.3\% across all evaluation metrics. We note that AugUndo improves VOICED and KBNet by larger amounts than FusionNet. For FusionNet, whether trained with conventional augmentation scheme or AugUndo, both seem to perform similarly. This is because FusionNet employs a learning-based densification network (ScaffNet), which is pretrained on synthetic datasets and frozen, that is used in both models. As the sparse depth maps in Waymo are much denser than KITTI, ScaffNet is able to approximate the dense depth map with small amounts of errors. This serves as an inductive bias for the downstream FusionNet, which learns the residual over the approximated depth map. As the reconstruction from ScaffNet exhibits high fidelity, the bias induced by ScaffNet on FusionNet causes FusionNet to perform only minor modifications to approximated depth map, leading to similar outputs whether trained with conventional augmentations or AugUndo. Nonetheless, we still observe consistent (albeit smaller) improvements when FusionNet is trained with AugUndo. 

For monocular depth estimation, we similarly evaluate Monodepth2, HR-Depth, and Lite-Mono models (trained on KITTI using standard augmentation pipelines and AugUndo) on Make3d. \tabref{tab:monocular_depth_prediction_generalization_outdoors} shows that models trained on KITTI with AugUndo generalizes well to Make3d, gaining an average of around 8\% improvement over all metrics and all models. Note: training Monodepth2 from their code repository reproduces their results on KITTI, but produces worse generalization results on Make3d than the weights they released; hence, we take the original weights released by the authors for evaluation. Nonetheless, we still improve over their best result.

\section{Sensitivity to Choice of Hyperparameters}
\label{sec:sensitivity_hyperparameters}
\begin{figure}[t]
    \centering
    \vspace{0em}
    \includegraphics[width=\linewidth]{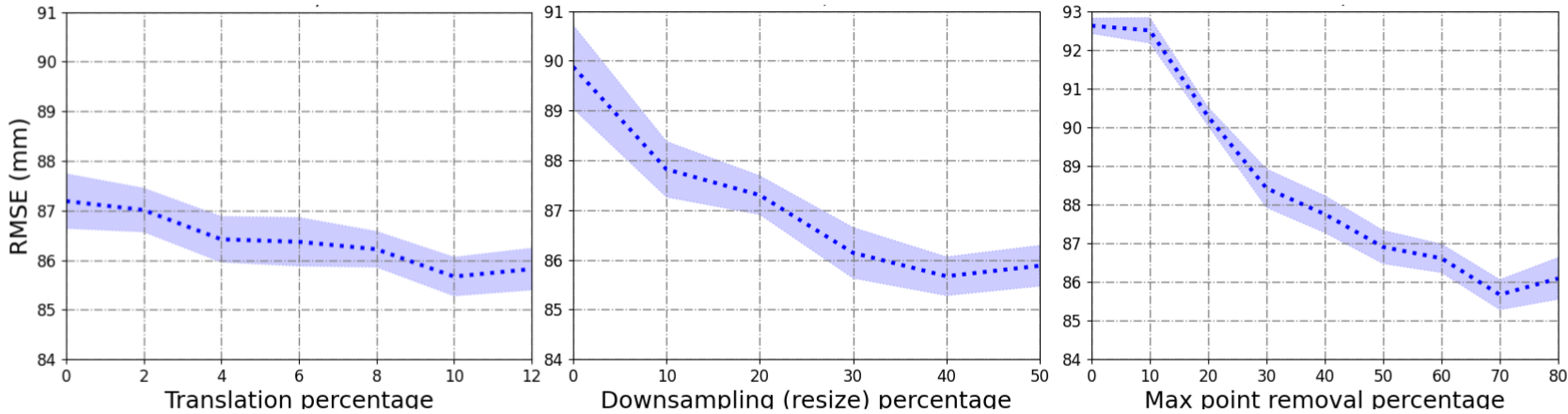}
    \vspace{-0em}
    \caption{\textit{Effect of different degree of augmentation on VOID1500.} }
    \label{fig:degree_aug}
\end{figure}

In the main paper and \secref{sec:implementation_details} above, we described in details of the set of augmentations and degrees of each augmentation used in our experiments to achieve the reported numbers. We note that the performance gains can be obtained by typical set and ranges of augmentations and does not require a meticulous selection of hyper-parameters. In the paper, we tried to push the limits and tested increasing degrees of augmentation until performance saturated. \figref{fig:degree_aug} shows the trends of improvement for resizing (down), translation, and points removal. Choosing even small degrees of augmentations yields some performance gain. We note that there are diminishing returns as we scale up augmentations to large amounts, i.e. 50\% or more in downsampling, which reduces the size of the observed image and predictions will have little to no detail. When computing the loss on these predictions on the original data, there is a considerable amount of noise and lends little to learning. Hence, at extreme degrees, we observe that performance eventually saturates. We also note that the choice of augmentation would depends on the task and dataset (i.e. vertical flip not applicable for KITTI since the camera is typical right-side-up for driving scenarios). 

\section{Additional Results for Depth Completion}
\label{sec:additional_results_depth_completion}
In the main paper, we demonstrated AugUndo for three unsupervised methods \cite{wong2020unsupervised,wong2021learning,wong2021unsupervised} on the VOID dataset. Due to space constraints, here, we show additional results on VOID1500 and VOID500 for a recent unsupervised distillation method, MonDi \cite{liu2022monitored} as well as a recent unsupervised depth completion method, DesNet \cite{yan2023desnet}, in \tabref{tab:depth_completion_void_mondi}. Additionally, we show MonDi and DesNet for VOID150 in \tabref{tab:depth_completion_void_sensitivity_sparsity}. We applied photometric transformations of random brightness from 0.5 to 1.5, contrast from 0.5 to 1.5, saturation from 0.5 to 1.5, and hue from -0.1 to 0.1. We applied image patch removal by selecting between 0.1\% to 0.5\% of the pixels and removing 5 $\times$ 5 patches centered on them -- approximately removing between 2.5\% to 12.5\% of the image. We applied random sparse depth point removal at a rate between 60\% and 95\% of all sparse points. We further applied geometric transformations of random horizontal and vertical flips, up to 10\% of translation, between -25 to 25 degrees of rotation, and random resize factor between 0.6 to 1.1.

\begin{table}[h]
    \centering
        \caption{
        \textit{Quantitative results of MonDi, a distillation-based unsupervised method, on VOID.} Reported scores are mean and standard deviation over four independent trials. 
        We evaluate MonDi trained on VOID1500, with standard augmentations and AugUndo, by testing them on VOID1500 and VOID500. Sparse depth maps within VOID1500 contains approximately 1500 points, and those within VOID500 contain approximately 3$\times$ less, respectively. AugUndo improves performance by an average of 5.33\% and 9.02\% across all methods and evaluation metrics on VOID1500 and VOID500, respectively.
    }
    \setlength\tabcolsep{1pt}
    \scalebox{0.73}{\begin{tabular}{l c c c c c c c c}
        \midrule
        & \multicolumn{4}{c}{VOID1500} & \multicolumn{4}{c}{VOID500} \\
        \cmidrule(lr){2-5} \cmidrule(lr){6-9}
        Method & MAE $\downarrow$& RMSE $\downarrow$& iMAE $\downarrow$& iRMSE $\downarrow$ & MAE $\downarrow$& RMSE $\downarrow$& iMAE $\downarrow$& iRMSE $\downarrow$ \\ 
        \midrule
        MonDi* 
        & 30.56$\pm$0.23 & 86.67$\pm$0.55 & 15.08$\pm$0.16 & 37.58$\pm$0.44 
        & 73.90$\pm$0.61 & 191.69$\pm$2.64 & 31.78$\pm$1.52 & 72.67$\pm$2.17  
         \\			
        + AugUndo
        & \textbf{29.02}$\pm$0.56 & \textbf{79.34}$\pm$1.17 & \textbf{14.56}$\pm$0.44 & \textbf{35.93}$\pm$1.12 
        & \textbf{69.90}$\pm$3.05 & \textbf{157.02}$\pm$7.46 & \textbf{29.61}$\pm$0.12 & \textbf{68.48}$\pm$0.57
        \\
        \midrule
        DesNet
        & 37.41$\pm$0.28 & 93.31$\pm$0.76 & 19.17$\pm$0.24 & 45.57$\pm$0.62
        & 74.89$\pm$0.67 & 170.32$\pm$2.03 & 35.62$\pm$0.72 & 78.30$\pm$1.31 
         \\			
        + AugUndo
        & \textbf{33.86}$\pm$0.60 & \textbf{86.05}$\pm$0.43 & \textbf{16.92}$\pm$0.29 & \textbf{41.25}$\pm$0.31 
        & \textbf{67.78}$\pm$1.10 & \textbf{153.46}$\pm$2.64 & \textbf{32.09}$\pm$0.48 & \textbf{71.96}$\pm$1.05
        \\
        \midrule
    \end{tabular}}
    \vspace{-0em}
\label{tab:depth_completion_void_mondi}
\end{table}

\textit{Results on MonDi and DesNet.} Overall, we improve MonDi by an average of 8.2\% across all evaluation metrics across all sparsity levels (VOID1500, VOID500, and VOID150). This is surprising as amongst all the tested methods, MonDi is the most light-weight, with only 5.3M parameters. As we reduce model capacity, one would expect that the network to saturate in the data variations that can be modeled. However, despite the size of network is much smaller (23.2\% less than KBNet), there is still a considerable gain when using AugUndo instead of their augmentation pipeline. This demonstrates efficacy and applicability of AugUndo; it can be used to improve methods with a range of capacities from tens of millions to several million. Also, we note that MonDi is an unsupervised distillation method (where it distills from unsupervised methods), so it is more closely related to supervised methods in supervision than unsupervised method. Even so, we observe a non-trivial improvement, which validates our discussion in Sec. 5 of the main paper regarding the applicability of AugUndo to \textit{supervised} methods.
Meanhwhile, we improves DesNet by an average of 10.64\% accross all evaluation metrics, highlighting again AugUndo's applicability to future unsupervised depth completion methods.

\begin{table}[t]
    \centering
        \caption{
        \textit{Quantitative results of KBNet with AugUndo as changes in camera pose or intrinsics (i.e. with and without depth adjustment, respectively) on VOID1500.}  With depth adjustments (change in camera pose), the performance of the model is slightly worse ($\approx$3.3\% in terms of percent gain) than without depth adjustments (change in camera intrinsics), yet still better than the baseline by a large margin. 
        }
    \scalebox{0.8}{ 
        \begin{tabular}{l c c c c c}
            \midrule
            Method & MAE $\downarrow$& RMSE $\downarrow$& iMAE $\downarrow$& iRMSE $\downarrow$ & Gain (\%)\\ 
            \midrule
            KBNet
            & 38.11$\pm$0.77 & 95.22$\pm$1.72 & 19.51$\pm$0.14 & 46.70$\pm$0.48 & - \\
            + AugUndo (intrinsics) 
            & \textbf{33.32}$\pm$0.18 & \textbf{85.67}$\pm$0.39 & \textbf{16.61}$\pm$0.29 & \textbf{41.24}$\pm$0.60 & \textbf{+12.29} \\
            + AugUndo (pose) 
            & 34.24$\pm$0.25 & 87.11$\pm$0.56 & 17.59$\pm$0.23 & 43.22$\pm$0.39 & +8.99 \\ 
            \midrule
        \end{tabular}
    }
    \vspace{-0em}
\label{tab:resize_comparison}
\end{table}

\textit{Modeling AugUndo as change in camera pose versus camera parameters.} In the main text, we discussed two ways of modeling our augmentation scheme: treating the augmented data as a result of changes in camera pose (i.e. motion) or in camera parameters (i.e. intrinsics). (1) Modeling the augmentation as camera motion requires adjusting sparse depth maps; for example, resizing can be treated as forward motion, so distance from camera to world surfaces need to be adjusted in the sparse depth map. Naturally, if choosing (1), then one would need to reproject the sparse depth points according to the camera motion. On the other hand, (2) modeling the augmentation as changes in the camera parameters does not require adjusting the sparse depth maps as the camera and 3D scene are both static; for example, resizing can now be treated as ``zooming in'' or an increase in focal length. 

\tabref{tab:resize_comparison} compares the two approaches. We test the resizing operation by adjusting sparse depth, and likewise, dense output depth. 
Specifically, we assume a pin-hole camera model. As perspective projection is a linear, we adjust the depth value by scaling them using the random scale factor recorded during the random resizing operation. 
We observe the following: Firstly, both methods of modeling improves KBNet on VOID: modeling as (1) improves KBNet by 8.99\% and (2) by 12.29\%, respectively. This verifies the efficacy of AugUndo under both modeling choice. Secondly, modeling AugUndo as changes in (2) camera parameters improves over (1) camera motion. Particularly, (2) yields an additional 3.3\% gain in the baseline and 3.6\% relative improvement over KBNet trained using (1). This justifies our choice of modeling AugUndo as changes in camera parameters.

\begin{table}[h!]
    \centering
        \caption{
        \textit{Naive geometric augmentations for unsupervised depth completion on VOID.} It is infeasible to conduct unsupervised training with naive geometric augmentations. Modifying geometric augmentations to ensure no borders are introduced allows the model to train, but performance degrades. ``+ Naive Geo Aug'' denotes models trained with naive geometric augmentations and ``+ Modified Geo Aug'' denotes mdoels trained with modified geometric augmentations.
        }
    \scalebox{0.8}{ 
        \begin{tabular}{l c c c c }
            \midrule
            Method & MAE $\downarrow$& RMSE $\downarrow$& iMAE $\downarrow$& iRMSE $\downarrow$ \\ 
            \midrule
            VOICED
            & 74.78$\pm$2.69 & 139.75$\pm$4.57 & 39.20$\pm$1.46 & 71.98$\pm$2.54 \\
            + Naive Geo Aug
            & 554.65$\pm$91.10 & 642.13$\pm$69.40 & 639.42$\pm$113.11 & 980.81$\pm$78.01 \\
            + Modified Geo Aug
            & 109.40$\pm$13.01 & 205.14$\pm$18.22 & 106.21$\pm$13.27 & 286.71$\pm$31.79 \\
            + AugUndo 
            & \textbf{52.73}$\pm$0.41 & \textbf{111.09}$\pm$0.92 & \textbf{26.93}$\pm$0.54 & \textbf{54.46}$\pm$0.38 \\
            \midrule
            FusionNet
            & 52.11$\pm$0.44 & 113.30$\pm$1.18 & 28.53$\pm$0.52 & 58.79$\pm$2.01 \\
            + Naive Geo Aug
            & 82.73$\pm$6.96 & 190.71$\pm$9.90 & 73.61$\pm$33.32 & 224.62$\pm$156.04 \\
            + Modified Geo Aug 
            & 57.72$\pm$6.67 & 138.52$\pm$15.87 & 31.91$\pm$5.19 & 74.41$\pm$14.64 \\
            + AugUndo
            & \textbf{41.16}$\pm$0.18 & \textbf{99.21}$\pm$0.39 & \textbf{22.23}$\pm$0.35 & \textbf{53.07}$\pm$1.30 \\
            \midrule
            KBNet
            & 38.11$\pm$0.77 & 95.22$\pm$1.72 & 19.51$\pm$0.14 & 46.70$\pm$0.48 \\
            + Naive Geo Aug
            & 1,065.4 $\pm$ 81.7  & 1,216.4 $\pm$ 54.4 & 5,693.5 $\pm$ 401.3 & 7,032.7 $\pm$ 283.0 \\
            + Modified Geo Aug
            & 51.77 $\pm$ 4.13 & 118.95 $\pm$ 7.67 & 27.66 $\pm$ 3.04 &  60.92 $\pm$ 5.62 \\
            + AugUndo 
            & \textbf{33.32}$\pm$0.18 & \textbf{85.67}$\pm$0.39 & \textbf{16.61}$\pm$0.29 & \textbf{41.24}$\pm$0.60 \\
            \midrule
        \end{tabular}
    }
    \vspace{-0em}
\label{tab:naive_augmentations_depth_completion}
\end{table}

\begin{table}[h!]
    \centering
        \caption{
        \textit{Naive geometric augmentations for unsupervised monocular depth estimation on KITTI.} It is infeasible to conduct unsupervised training when naively applying geometric augmentations. Modifying geometric augmentations to ensure no borders are introduced allows the model to train, but performance degrades. ``+ Naive Geo Aug'' denotes models trained with naive geometric augmentations and ``+ Modified Geo Aug'' denotes mdoels trained with modified geometric augmentations.
        }
    \scalebox{0.8}{ 
        \begin{tabular}{l c c c c c c c}
            \midrule
            Method  & RMSE $\downarrow$& Abs Rel $\downarrow$& Sq Rel $\downarrow$& $\delta < 1.25 \uparrow$ & $\delta < 1.25^2\uparrow$ & $\delta < 1.25^3\uparrow$ \\ 
            \midrule
            Monodepth2
            & 4.794 $\pm$ 0.035 & 0.117$\pm$0.001 & 0.845$\pm$ 0.030& 0.869$\pm$ 0.004& 0.959$\pm$ 0.001& 0.982$\pm$0.001 \\
            + Naive Geo Aug
            & 6.612$\pm$0.18 & 0.174$\pm$0.009& 2.057$\pm$0.32& 0.794$\pm$0.02 & 0.932$\pm$0.004 & 0.971$\pm$0.002 \\
            + Modified Geo Aug 
            & 4.911$\pm$0.046 & 0.119$\pm$0.002 & 0.926$\pm$0.023 & 0.870$\pm$0.004 & 0.958$\pm$0.001 & 0.981$\pm$0.001 \\
            + AugUndo 
            & \textbf{4.739}$\pm$0.032 & \textbf{0.113}$\pm$0.000 & \textbf{0.862}$\pm$0.030 & \textbf{0.879}$\pm$0.002 & \textbf{0.960}$\pm$0.001 & \textbf{0.982}$\pm$0.001 \\

            \midrule
            HR-Depth
		
            & 4.626 $\pm$ 0.032 & 0.113$\pm$0.001 & 0.797$\pm$ 0.022& 0.879$\pm$0.002& 0.961$\pm$ 0.001& 0.982$\pm$0.000 \\
            + Naive Geo Aug
            & 6.06$\pm$0.022 & 0.159$\pm$0.003& 1.40$\pm$0.016& 0.795$\pm$0.010 & 0.939$\pm$0.003 & 0.979$\pm$0.002 \\
            + Modified Geo Aug
            &  4.718$\pm$0.018 &0.116$\pm$0.001 & 0.844$\pm$0.005	 & 0.875$\pm$0.001 &0.960$\pm$0.001	& 0.982$\pm$0.001\\
            + AugUndo 
            & \textbf{4.610}$\pm$0.029 & \textbf{0.111}$\pm$0.001 & \textbf{0.794}$\pm$0.021& \textbf{0.883}$\pm$0.001 & \textbf{0.962}$\pm$0.001 & \textbf{0.983}$\pm$0.001 \\
            \midrule
        \end{tabular}
    }
    \vspace{-0em}
\label{tab:naive_augmentations_depth_estimation}
\end{table}

\textit{Naive geometric augmentations.} In the main paper, we discussed that naively applying geometric augmentations can be detrimental to model performance and even prevent one from training them. As unsupervised depth completion and unsupervised monocular depth estimation assume rigid motion within the image triplet comprising of a training example, geometric augmentations that introduce some form of border padding (e.g., translation, rotation) will yield constant or edge extended borders across images (i.e., no motion in those regions). The lack of motion will result in PoseNet predicting near identity pose. Hence, naively incorporating geometric augmentations will prevent the model from properly learning depth and pose. This is demonstrated in \tabref{tab:naive_augmentations_depth_completion} and \ref{tab:naive_augmentations_depth_estimation} for rows marked with ``+ Naive Geo Aug''. Nonetheless, one can make modifications (no translation, center cropping on rotation, and resizing to the same shape for a batch) to ensure no borders are introduced during augmentation. Yet, we do not perform the proposed ``undo-ing'' process. While this will allow the model to train, however, the effects of artifacts, loss during resampling, color distortion, and intensity saturation will lead to performance degradations worse than the baseline. This is shown in \tabref{tab:naive_augmentations_depth_completion} and \ref{tab:naive_augmentations_depth_estimation} for rows marked with ``+ Modified Geo Aug''.

\section{Additional Results on Monocular Depth Estimation}
\label{sec:additional_results_monocular_depth}

\begin{figure*}[h]
    \centering
    \includegraphics[width=1\textwidth]{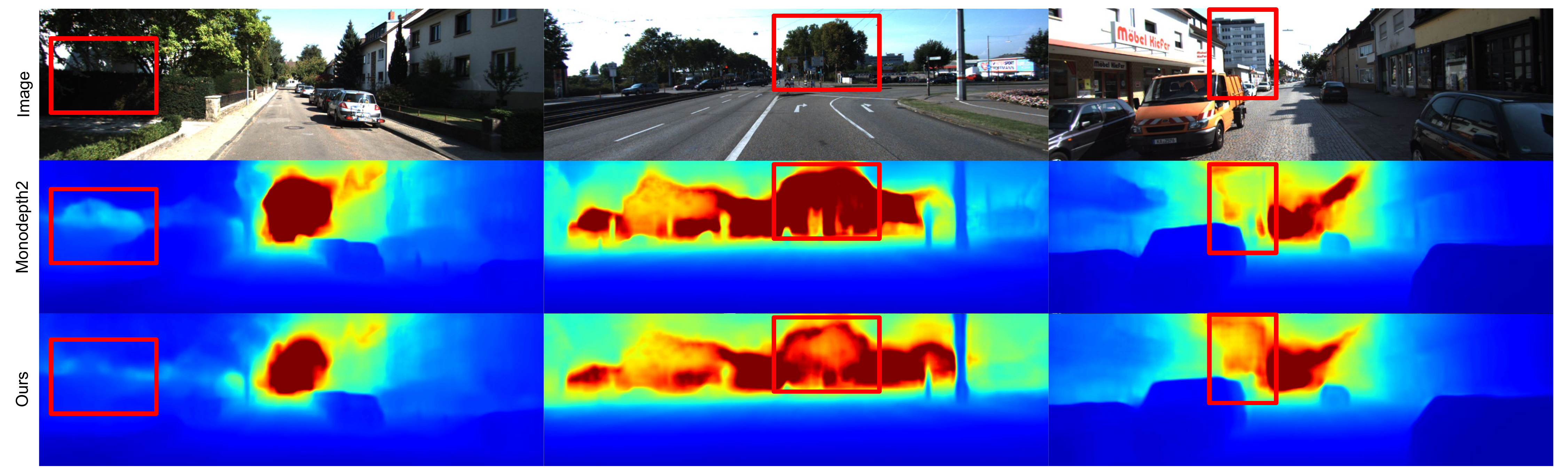}
    \caption{Qualitative result of MonoDepth2 on KITTI. Red bounding boxes highlight areas where training with our augmentation scheme improves Monodepth2, i.e., wall and vegetation on left, trees in middle and building on right.}
    \label{fig:depth_estimation_kitti}
\end{figure*}

Applying our methods yields qualitative improvements in the depth prediction. In \figref{fig:depth_estimation_kitti}, we can see that our method better captures the building in the last column. Similarly, we observe similar improvements in the middle column for the trees that are located at a distance. This is in part thanks to augmentations such as resizing so that we can simulate objects at far and close distances. Additionally, training with AugUndo also improves over the baseline method on ambiguous regions such as the wall and vegetation in the first column, despite using the same hyper-parameter during training with the exception of data augmentation.

\section{Limitations}
\label{sec:limitations}
While we have proposed an algorithm to scale up augmentations for depth completion, a multimodal 3D reconstruction problem, admittedly our augmentations are limited to 2D. In the case of this inverse problem, the  nuisance variability simulated by AugUndo are, in fact, projections of those in the 3D scene, which we do not model directly. Additionally, augmentations are used in other vision tasks including multiview stereo, binocular stereo, optical flow, etc. While we have shown that AugUndo can be applied to unsupervised and distillation (with dense supervision from pseudo ground truth), we have limited the scope of our method for depth completion, which considers a single image and sparse depth map as input. We foresee that AugUndo can also incorporate other modalities, such as tactile \cite{yang2024binding}, be extended to other settings \cite{park2024test}, and tasks, including deep feature visualization \cite{lao2024sub}, semantic segmentation \cite{chen2017deeplab,han2023spatial,lao2024depth,liu2019auto,strudel2021segmenter,wong2021small}, and object detection \cite{carion2020end,lao2017minimum,lao2019minimum,redmon2016you,tan2020efficientdet}. Additionally, we hypothesize that AugUndo can also be extended towards multi-frame geometric tasks such as stereo \cite{berger2022stereoscopic,wong2021stereopagnosia,xu2020aanet,wang2021patchmatchnet}, optical flow \cite{lao2024diffeomorphic,lao2018extending,sun2018pwc,teed2020raft,zhang2024adaptive,zhang2024heteroscedastic}, etc., but one must account for their specifics and problem setups, i.e. stereo assumes frontoparallel views. Lastly, like all scaling problems, AugUndo is eventually limited by diminishing returns. As certain augmentation are pushed to extremes, i.e., maximum brightness such that the image is ``white'', large spatial reduction such that the image is small, there is little to no information in the input to infer depth; even if the supervision signal exists, it would be mapping nonsensical inputs to depth maps, which does not improve performance (as illustrated by \figref{fig:degree_aug}), and lending to  saturation in performance gain.

\end{document}